\documentclass[10pt]{article} %
\usepackage[accepted]{tmlr}

\usepackage{amsmath,amsfonts,bm}

\def\Figref#1{Figure~\ref{#1}}

\def\Secref#1{Section~\ref{#1}}

\def\eqref#1{equation~\ref{#1}}

\def\1{\bm{1}}

\def\vtheta{{\bm{\theta}}}

\def\vx{{\bm{x}}}

\def\vz{{\bm{z}}}

\DeclareMathAlphabet{\mathsfit}{\encodingdefault}{\sfdefault}{m}{sl}
\SetMathAlphabet{\mathsfit}{bold}{\encodingdefault}{\sfdefault}{bx}{n}

\DeclareMathOperator*{\argmax}{arg\,max}

\usepackage{hyperref}
\usepackage{url}
\usepackage{amsmath}
\usepackage{subcaption}
\usepackage{graphicx}
\usepackage{booktabs}
\usepackage{algorithm}
\usepackage{algpseudocode}

\usepackage{xcolor, soul}         %
\sethlcolor{pink}
\usepackage{graphicx}
\usepackage{titletoc} %
\usepackage{wrapfig}
\usepackage{float}

\makeatletter
\newcommand{\appendixtoccolor}{%
  \begingroup
  \hypersetup{linkcolor=black}%
}
\newcommand{\restorelinkcolor}{%
  \endgroup
}
\makeatother

\title{Mixed neural posterior estimation for simulators with\\discrete and continuous parameters}

\author{
\hspace{-0.14cm}Jan Boelts$^{1,\dagger}$,
Cornelius Schröder$^{2,3,4,\dagger}$,
Jonas Beck$^{2,3,4,5,\dagger}$,\\
Jakob H.~Macke$^{2,3,4,\ddag}$,
Michael Deistler$^{3,4,6,\ddag}$,
Daniel Gedon$^{2,3,4,\ddag}$ \\
\\
{\normalfont
$^1$ appliedAI Institute for Europe \\
$^2$ ELLIS Institute Tübingen  \\
$^3$ Machine Learning in Science, University of Tübingen \\
$^4$ Tübingen AI Center, University of Tübingen  \\
$^5$ Hertie Institute for AI in Brain Health, University of Tübingen \\
$^6$ Max Planck Institute for Biological Intelligence \\
$^\dagger$ Equal contribution, $^\ddag$ Joint supervision
}
}

\def\month{09}  %
\def\year{2026} %
\def\openreview{\url{https://openreview.net/forum?id=fOZZeRfdrC}}

\begin{document}

\maketitle

\begin{abstract}
Neural Posterior Estimation (NPE) enables rapid parameter inference for complex simulators with intractable likelihoods. NPE trains an inference network to estimate a probability density over parameters given data, typically assumed to be \emph{continuous}. However, many scientific models involve parameter spaces that are \emph{mixed}, that is, they contain both discrete and continuous dimensions.
We address this limitation by extending NPE to mixed parameter spaces through an inference network that jointly handles discrete and continuous parameters.
The inference network factorizes the joint posterior into discrete and continuous components, combining an autoregressive classifier for the discrete parameters with a generative model for the continuous parameters, trained jointly under a single simulation-based objective. 
In addition, we propose a diagnostic tool to assess the calibration of the mixed posterior approximation.
Across tractable toy examples and real-world scientific simulators, our joint inference approach yields accurate and calibrated posteriors. 
The inference framework is available in the \texttt{sbi} Python package.
\end{abstract}

\section{Introduction}
Neural Posterior Estimation (NPE) has become a central tool for simulation-based inference (SBI) \citep{cranmer2020frontier, deistler2025simulation}. 
NPE enables posterior inference in simulator models using only simulations from the forward model, without requiring access to likelihood evaluations.
This setting arises across the natural sciences, cognitive science, epidemiology, and engineering \citep{mckinley2014simulation,goncalves2020training,fengler2021LAN}. In addition, NPE amortizes the cost of simulation and training and rapidly performs inference for any observation, enabling real-time and high-throughput applications \citep{dax2021real, von2022mental}. Over the past years, NPE has seen broad adoption, with software libraries making it accessible to domain scientists with little tuning or customization needed \citep{boeltsdeistler2025reloaded,kuhmichel2026bayesflow}. As these libraries lower the barrier to adoption, NPE is increasingly applied directly to simulators as they arise in practice.

A fundamental challenge in using NPE is that many simulators involve both continuous and discrete parameters (i.e., categorical variables). We refer to these simulators as mixed (discrete--continuous). Discrete parameters often arise naturally: phenomena such as ion channel type in neuroscience \citep{schroeder2024simultaneous}, change-point locations in time-series simulators \citep{adams2007bayesian,altamirano2023robust}, switching dynamical systems \citep{linderman17a,fu2024simultaneous}, queueing simulators with discrete service regimes \citep{gross_2008}, or embedded model selection variables \citep{radev2021amortized,gloeckler2026scalable} are inherently discrete. 
Handling such mixed parameter spaces is non-trivial for NPE. The joint posterior over mixed parameters is a hybrid object: a mixture of continuous densities, one per discrete configuration. 
Standard continuous inference networks based on normalizing flows or diffusion-based approaches---as commonly used in NPE---cannot naturally represent this.
Even if simulator likelihoods are available and standard inference methods like MCMC can be used for mixed parameters, those need to be carefully tuned and they do not benefit from amortization across observations. 
Thus, despite their ubiquity, simulators with mixed parameters remain insufficiently supported by current SBI methods and toolboxes.
Recent work has begun to address this gap from different angles: diffusion-based SBI methods embed discrete parameters in continuous spaces or treat them as model indices \citep{ghiglino2026diffusion_mixed_estimation, schroeder2024simultaneous, gloeckler2026scalable}, while mixed MCMC approaches extend Hamiltonian dynamics to handle discrete variables alongside continuous ones \citep{nishimura2020discontinuousHMC, zhou2020mixed}. We discuss these approaches and their trade-offs in detail in Section~\ref{sec:discussion}.

\begin{figure}[t]
    \centering
    \includegraphics[width=1.0\linewidth]{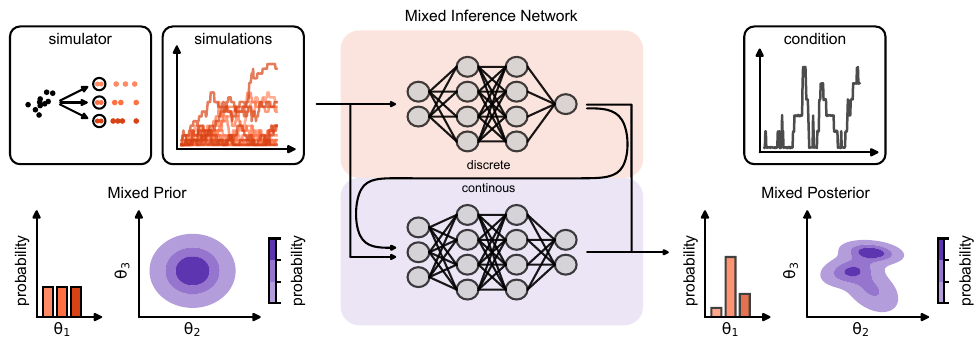}
    \caption{\textbf{Inference method overview.}
    MNPE is trained on a dataset of parameter--simulation pairs~$\{(\vtheta,\vx)\}$ where $\vtheta$ is sampled from a mixed prior over discrete and continuous parameters. The inference network consists of a subnetwork for the discrete dimensions (orange), implemented as a masked autoregressive density estimator (MADE), and one for the continuous dimensions (violet), which can be a standard inference network for continuous posterior estimation (e.g., normalizing flow, diffusion model). At inference time, the network is conditioned on an observation~$\vx_o$ and provides the joint posterior over discrete and continuous dimensions.  
    }
    \label{fig:graph_abstract}
\end{figure}

Here, we extend NPE to mixed parameter spaces by introducing an inference network that handles a fixed number of discrete and continuous parameters jointly (\Figref{fig:graph_abstract}). We refer to this extension as Mixed Neural Posterior Estimation (MNPE). 
Our inference network handles multiple discrete parameters with differing class counts, and is compatible with any continuous generative network, including normalizing flows and diffusion models. 
In addition, we show how to assess calibration of mixed posteriors by combining simulation-based calibration \citep{talts2018validating} for continuous parameters with expected calibration error \citep{guo2017calibration} for discrete parameters, and propose an empirical finite-sample baseline for discrete calibration.
We demonstrate that MNPE performs accurate inference on several problems of increasing difficulty: We first verify that our method works on a tractable mixed Gaussian simulator with an analytical reference solution. We then show that it matches MCMC on a queueing simulator with available likelihoods. Finally, we show that MNPE obtains well-calibrated posteriors for an intractable biophysical neuroscience simulator. 
MNPE is implemented in the publicly available \texttt{sbi} toolbox \citep{boeltsdeistler2025reloaded} with code and documentation are available at \url{https://sbi.readthedocs.io/}.

\section{Background}

\subsection{Neural Posterior Estimation}

Neural Posterior Estimation (NPE) is a powerful method for simulation-based inference \citep{papamakarios2016fast,greenberg2019automatic}. NPE first generates a dataset of parameter--simulation pairs~$(\vtheta,\vx)$ by sampling from the prior~$p(\vtheta)$ and running the simulator~$\vx \sim p(\vx \mid \vtheta)$. It then trains an inference network~$q(\vtheta \mid \vx)$ (i.e., a conditional generative model) to directly approximate the posterior~$p(\vtheta \mid \vx)$. After training, the inference network can be queried at any observation~$\vx_o$ and can directly draw posterior samples, making NPE an amortized method where a single training run supports repeated inference over different observations at negligible cost.

Standard NPE inference networks are designed for continuous parameters only. They parameterize the posterior as a continuous probability density, typically via a normalizing flow or diffusion model defined on~$\mathbb{R}^k$ \citep{greenberg2019automatic, radev2020bayesflow, simons2023neural}. Such networks are well-suited for smooth posteriors supported on open domains, but are not applicable when the parameter space is discrete or mixed discrete--continuous. That is, when $\vtheta=(\vtheta_d,\vtheta_c)$, with $l$ discrete dimensions $\vtheta_d\in\mathcal{D} = D_1\times \dots\times D_l$, where ${D}_i$ is a finite set for each discrete dimension $i$, and $k$ continuous dimensions, $\vtheta_c\in\mathbb{R}^k$.

\subsection{Autoregressive models}

Any multi-dimensional probability distribution can be factorized as the product of its conditionals
\begin{equation*}
    p(\vz) = p(z_1, ..., z_M) = p(z_1) \cdot p(z_2|z_1) \cdot \;... \;\cdot \; p(z_M|z_{1,...,M-1}),
\end{equation*}
with $N$ training samples and $i$ indexes the training sample (omitted from all $z$ to simplify notation).
Autoregressive models use this factorization to cast the estimation of a high-dimensional probability distribution~$p(\vz)$ into a series of one-dimensional estimation problems. Masked autoregressive density estimators (MADEs) define a feedforward network~$q(\vz)$ which take samples as input and return, for every dimension, values that parameterize each conditional probability distribution (i.e., $p(z_1)$, $p(z_2|z_1)$,...). MADEs mask individual weights to enforce the autoregressive property (e.g., $p(z_2 | z_1)$ should not depend on $z_{>1}$). With this setup, the joint log-probability can be evaluated in parallel across all dimensions, which enables efficient training with the negative log-likelihood
\begin{equation*}
    \mathcal{L} = -\frac{1}{N} \sum_i^N \log q(\vz) = -\frac{1}{N} \sum_i^N \left( \log q(z_1) + \log q(z_2|z_1) + \;... \; + \; \log q(z_M|z_{1,...,M-1})\right),
\end{equation*}
where $N$ is the number of training samples and $i$ indexes the training sample. After training, one can draw samples from the MADE by sequentially performing $M$ forward passes (i.e., one forward pass per dimension).

We will use a conditional autoregressive model to estimate the distribution of parameters $\vtheta$ given simulation outputs $\vx$. As such, the distribution will be over $\vtheta$ (instead of $z$), and all distributions will additionally be conditioned on $\vx$.

\section{Methods}
\label{sec:methods}

\subsection{Inference for mixed parameter spaces}

We factorize the posterior over $\vtheta=(\vtheta_d,\vtheta_c)$ as
\begin{equation*}
    p(\vtheta_d,\vtheta_c \mid \vx) = p(\vtheta_d \mid \vx)\, p(\vtheta_c \mid \vtheta_d, \vx).
\end{equation*}
This factorization separates the inference problem into two sub-problems: classification over the discrete space $\mathcal{D}$, and conditional density estimation in a purely continuous domain, given the discrete state. We choose this factorization because $\vtheta_d$ is typically low-dimensional. The reverse factorization would instead require the continuous estimator to model the potentially multimodal marginal $p(\vtheta_c \mid \vx)$ directly, without the benefit of conditioning on the discrete state. We compare both factorization orders, as well as a variant that drops the conditioning entirely (Appendix~\ref{app:factorization}).

This naturally gives rise to two dedicated inference networks, which together define the \emph{Mixed Neural Posterior Estimation} (MNPE) algorithm. For the discrete factor $p(\vtheta_d \mid \vx)$, we use an autoregressive network, specifically a Masked Autoregressive Density Estimator (MADE; \citealt{germain2015made}), which returns the class probabilities for each discrete parameter dimension conditioned on all preceding ones. Therefore, a MADE can handle multiple discrete parameters, each with a varying number of classes $|D_i|$. When $\vtheta_d$ contains multiple discrete dimensions, sampling from the discrete density estimator is performed autoregressively according to the ordering of the categorical MADE. The ordering of the discrete dimensions in this factorization is arbitrary, and we find that swapping it does not affect posterior accuracy (Appendix~\ref{app:ordering}).
For the continuous factor $p(\vtheta_c \mid \vtheta_d, \vx)$, we use a standard conditional generative model (a normalizing flow or diffusion model) conditioned on both $\vtheta_d$ and $\vx$. This modular design allows any continuous inference network to be substituted without modifying the discrete component.
As with other NPE inference methods, MNPE avoids the need for specialized MCMC kernels or sampler tuning as it targets directly the posterior. 

We train our mixed inference network $q(\vtheta_d, \vtheta_c \mid \vx)$ using the joint negative log-probability of mixed and discrete parameters $-\log q(\vtheta_d, \vtheta_c \mid \vx)$. Using the factorization defined above, the MNPE training objective factorizes as
\begin{equation*}
    -\log q(\vtheta_d, \vtheta_c \mid \vx) = -\log q(\vtheta_c \mid \vtheta_d, \vx) -\log q(\vtheta_d \mid \vx),
\end{equation*}
which separates the loss into a loss over continuous and discrete parameters. For the continuous part, we use a normalizing flow with parameters $\phi_c$ and directly evaluate $\log q_{\phi_c}(\vtheta_c \mid \vtheta_d, \vx)$. For the discrete parameters, the negative log-likelihood reduces to a cross entropy loss over the parameters $\phi_d$ of the discrete estimator. 
While the disjoint networks permit optimizing them separately, we train them in a single procedure so that the two factors are fitted and evaluated as one joint posterior estimator. 
Note that the two subnetworks share no parameters, so the loss gradients do not interact, i.e., the categorical loss leaves the continuous subnetwork unchanged, and vice versa. 
Joint training is therefore a matter of convenience rather than a requirement, and we confirm empirically that it matches separate training (Appendix~\ref{app:jointsep}).

\begin{wrapfigure}{r}{0.54\textwidth}
\centering
\vspace{-22pt}
\begin{minipage}{0.9\linewidth}
\begin{algorithm}[H]
\caption{Sampling from the MNPE posterior}
\label{alg:mnpe-sampling}
\begin{algorithmic}[1]
\small
\Require Observation $\vx_o$, number of samples $S$
\Ensure Joint posterior samples
$\{(\vtheta_d^{(s)},\vtheta_c^{(s)})\}_{s=1}^{S}$

\For{$s=1,\ldots,S$}
    \State Sample
    $\vtheta_d^{(s)}
    \sim q_{\phi_d}(\vtheta_d \mid \vx_o)$
    \State Sample
    $\vtheta_c^{(s)}
    \sim q_{\phi_c}
    (\vtheta_c \mid \vtheta_d^{(s)},\vx_o)$
\EndFor

\State \Return
$\{(\vtheta_d^{(s)},\vtheta_c^{(s)})\}_{s=1}^{S}$
\end{algorithmic}
\end{algorithm}
\end{minipage}
\end{wrapfigure}
At test time, samples are drawn ancestrally according to the posterior factorization (Algorithm~\ref{alg:mnpe-sampling}). We first sample a discrete parameter vector from the discrete posterior. 
The sampled discrete vector is then passed, together with the observation, as a conditioning input to the continuous estimator. Each continuous sample is therefore conditioned on its corresponding discrete sample, rather than on the categorical probabilities or on an average over discrete states.
Consequently, an error in the discrete posterior re-weights the mixture of per-state conditionals rather than distorting them, which increases the error of the continuous posterior at most linearly (Appendix~\ref{app:propagation}).

Finally, we note that the factorization of the loss would also permit other loss functions and, thus, other architectures for the discrete and continuous parameters (e.g., flow-matching or diffusion models for the continuous parameters).

\subsection{Calibration for mixed posteriors}
\label{subsec:calibration}

In the absence of a reference posterior, statistical calibration provides a principled way to assess posterior quality on average: a calibrated posterior requires that events assigned probability $p$ occur with frequency $p$ under the ground truth posterior \citep{gneiting2007probabilistic}. 
Standard calibration tools in SBI, such as simulation-based calibration (SBC) \citep{cook2006validation,talts2018validating} allow checking posterior calibration without requiring direct access to the underlying reference posteriors. However, they assume continuous parameters and do not directly apply to mixed posteriors. We suggest combining rank-based calibration techniques like SBC for the continuous dimensions with established calibration methods from the classification literature for the discrete dimensions.

\paragraph{Continuous parameters.}
SBC generates a calibration set of parameter-data pairs $(\vtheta_i, \vx_i) \sim p(\vtheta)\,p(\vx \mid \vtheta)$ from the prior and the simulator and obtains approximate posterior samples $\{\vtheta^{(s)}\}_{s=1}^S \sim q(\vtheta \mid \vx_i)$ for each. 
For a specified scalar projection $f: \vtheta \rightarrow \mathbb{R}$, SBC then checks whether the rank of $f(\vtheta)$ (i.e., the true parameters) is uniformly distributed within $f(\vtheta^{(s)})$ (i.e., posterior samples). This is a necessary condition for a well-calibrated posterior, i.e., deviation from uniform ranks implies miscalibration in the posterior estimate. A common choice for $f(\cdot)$ is the projection to single dimensions independently, computing a marginal rank statistic per dimension (see also Appendix~\ref{app:alternative_calibration}). 
Deviations of the distribution over ranks from uniformity are visualized as rank histograms per parameter dimension, or as a cumulative distribution function (CDF) of the empirical rank. 
We can summarize the deviation from optimality by the error over diagonal (EoD), defined as the mean absolute error of the empirical rank CDF from the diagonal, averaged across dimensions.

\paragraph{Discrete parameters.}
For discrete parameters, rank-based SBC is not applicable because the posterior is a probability mass function rather than a continuous density \citep{talts2018validating}. 
Therefore, we instead use reliability diagrams and the top-label expected calibration error (ECE; \citealt{guo2017calibration}): 
For each discrete dimension $i$ and each test pair
$(\vtheta, \vx)$, we define the predicted class and associated confidence as
\begin{equation*}
    \hat{\theta}_{d_i} = \argmax_j\, q(\theta_{d_i}{=}j \mid \vx), \qquad \hat{p}_{d_i} = q(\theta_{d_i}{=}\hat{\theta}_{d_i} \mid \vx),
\end{equation*}
and bin the $N$ test pairs by confidence into $B$ equal-width bins. For a perfectly calibrated posterior, the empirically observed class accuracy matches the confidence of the posterior probability across all bins, i.e., $\operatorname{acc}(b) = \operatorname{conf}(b)$ for every bin $b$, where
\begin{equation*}
    \operatorname{acc}(b) = \frac{1}{n_b}\sum_{j \in b} \mathbf{1}(\hat{\theta}_{d_i,j} = \theta_{d_i,j}), \qquad \operatorname{conf}(b) = \frac{1}{n_b}\sum_{j \in b} \hat{p}_{d_i,j},
\end{equation*}
with $n_b$ the number of test pairs in bin $b$. The ECE summarizes per-bin deviations as a scalar,
\begin{equation*}
    \text{ECE}(d_i) = \sum_{b=1}^{B} \frac{n_b}{N}\,\bigl|\,\operatorname{acc}(b) - \operatorname{conf}(b)\,\bigr|.
\end{equation*}
Reliability diagrams, typically shown as bar plots of $\operatorname{acc}(b)$ against
$\operatorname{conf}(b)$, provide a calibration diagnostic for discrete parameters
analogous to SBC rank CDFs for continuous ones: in both cases, agreement with the
diagonal indicates calibration.

\paragraph{Interpreting calibration results.}
Together, marginal SBC for continuous $\vtheta_c$ and marginal ECE for discrete $\vtheta_d$ provide complementary calibration checks for all dimensions of a mixed posterior. 
These diagnostics can be interpreted both visually and quantitatively. 
A well-calibrated posterior yields uniform rank CDFs and diagonal reliability diagrams, with EoD and ECE close to zero. 
Visually, specific shapes in the SBC rank histograms reveal miscalibration patterns, e.g., u-shaped histograms or above diagonal CDFs indicate overconfidence \citep{talts2018validating}; 
for reliability diagrams, bars below the diagonal (confidence exceeds accuracy) indicate overconfidence, while bars above indicate underconfidence.

Quantitatively, interpreting EoD and ECE values requires finite-sample baselines, since both metrics remain strictly positive even under perfect calibration due to finite sample noise. 
For the EoD, the expected deviation under uniform ranks is known analytically, independent of the posterior estimator, and provides a baseline with an associated confidence band \citep{talts2018validating}. 
For the ECE, no such posterior-independent baseline exists: the finite-sample noise floor depends on how test cases are distributed across confidence bins, which is determined by the posterior estimator itself and therefore varies with training budget. 
Therefore, to distinguish actual miscalibration in discrete posterior dimensions from statistical noise, we propose an empirical baseline under the assumption of a perfectly calibrated classifier (Appendix~\ref{app:ece_baseline}).

\section{Results}

We validate our approach on three simulators with increasing complexity. First, we showcase MNPE on a mixed Gaussian simulator which is analytically tractable and allows for an analytical reference solution. Second, we apply MNPE to a queueing simulator with known likelihood, and compare it to a numerically exact reference solution (additional comparison to a common mixed MCMC task in Appendix~\ref{app:coal}). In the last example we turn to a `black-box' simulator with intractable likelihood, the Hodgkin--Huxley simulator. 

We use three complementary metrics to measure the quality of the approximated posteriors. 
When reference posteriors are available, we use the \textit{Classifier two-sample test} (C2ST; \citealt{friedman2004multivariate,lopez-paz2017revisiting,lueckmann2021benchmarking}), which trains a classifier to distinguish MNPE posterior samples from a reference distribution. A score near 0.5 (chance level) indicates the two distributions are indistinguishable. 
To test statistical calibration in the absence of a reference, we use SBC for continuous posterior dimensions and ECE for discrete parameters.

\subsection{Gaussian example}
\label{sec:toy}

\begin{figure}[t]
    \centering
    \includegraphics[width=\textwidth]{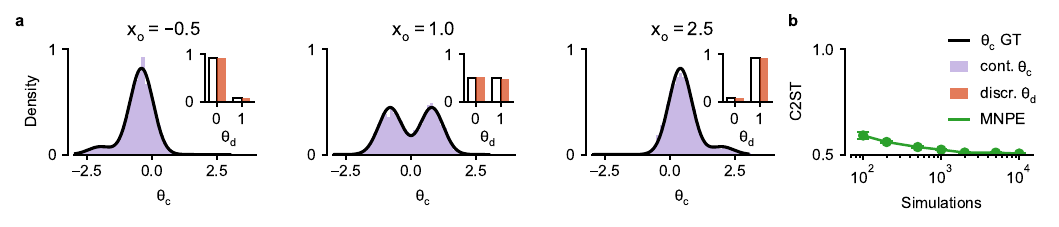}
    \caption{\textbf{Tractable Gaussian benchmark.}
    \textbf{(a)}~MNPE posteriors for three observations spanning different regimes: $x_o = -0.5$ (discrete state $\vtheta_d{=}0$ dominates, unimodal), $x_o = 1.0$ (uncertain $\vtheta_d$, bimodal), and $x_o = 2.5$ ($\vtheta_d{=}1$ dominates). Histograms show the continuous marginal posterior $p(\theta_c \mid x_o)$ from MNPE samples (violet), with the analytical ground-truth density overlaid (black). Inset bar charts compare the discrete posterior $P(\vtheta_d \mid x_o)$ between GT (black outline) and MNPE (orange). 
    \textbf{(b)}~C2ST on the joint $(\theta_c, \vtheta_d)$ distribution as a function of training simulations. Error bars show standard deviations over five independent training runs, each evaluated on ten held-out test observations. The score approaches the chance level of $0.5$ with ${\sim}1{,}000$ simulations.
    }
    \label{fig:toy}
\end{figure}

We construct a tractable toy example with one continuous and one discrete latent variable coupled through a shared Gaussian observation
\begin{align*}
\begin{split}
    \vtheta_c &\sim \mathcal{N}(0, 1), \\
    \vtheta_d &\sim \mathrm{Bernoulli}(0.5), \\
    x \mid \vtheta_c, \vtheta_d &\sim \mathcal{N}(\vtheta_c + a\,\vtheta_d,\; \sigma^2),
\end{split}
\end{align*}
with shift $a = 2$ and noise $\sigma = 0.5$. The discrete parameter $\vtheta_d$ shifts the observation mean, making the marginal posterior $p(\vtheta_c \mid x)$ a two-component Gaussian mixture whose weights depend on~$x$. Specifically, $p(\vtheta_d = 1 \mid x)$ follows a logistic form and the conditional $p(\vtheta_c \mid \vtheta_d, x)$ is Gaussian with closed-form mean and variance, yielding an analytical reference posterior. This example can be interpreted as a noisy channel simulator, where the receiver must jointly infer the transmitted signal $\vtheta_c$ and the channel mode $\vtheta_d$ from a single noisy observation. This tractable toy example captures the basic challenge of mixed-parameter inference: discrete-continuous coupling that induces multimodal continuous posteriors.

We train MNPE on increasing simulation budgets $N = 100, \ldots 10{,}000$ using a neural spline flow for the continuous parameters and a categorical MADE for the discrete parameters. We compute MNPE posteriors for three test observations spanning different regimes of the discrete variable (\Figref{fig:toy}a): $\vtheta_d = 0$ dominates for small $x$, $\vtheta_d = 1$ dominates for large $x$, and the posterior is bimodal at intermediate values. MNPE accurately recovers the ground-truth posterior in all cases, including the bimodal regime where both discrete states are plausible ($x_o = 1.0$). The inset bar charts confirm that the discrete marginal $P(\vtheta_d \mid x_o)$
is well calibrated across regimes (\Figref{fig:toy}a). We additionally compute the C2ST score to the ground truth posterior, based on $1{,}000$ posterior samples. The C2ST score converges to the chance level of $0.5$ at around ${\sim}1{,}000$ training simulations, indicating that MNPE provides accurate posterior estimates (\Figref{fig:toy}b).

\subsection{Tandem queueing simulator}
\label{sec:queue}

\begin{figure}[t]
    \centering
    \includegraphics[width=\textwidth]{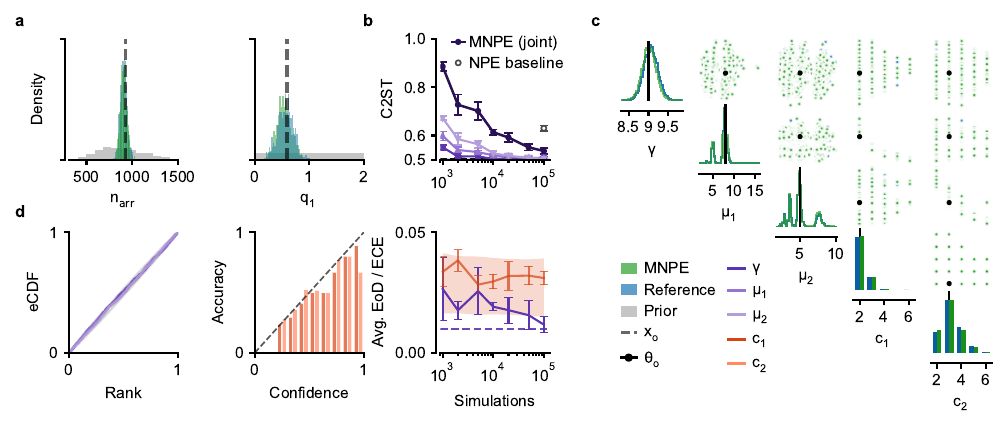}
    \caption{\textbf{Tandem queueing simulator}.
    \textbf{(a)}~Posterior predictive simulations for arrival counts $n_{\mathrm{arr}}$ and queue length $q_1$, comparing MNPE (green), MCMC (blue), and prior predictive (grey). Dashed line marks~$x_o$.
    \textbf{(b)}~C2ST for joint (grey) and marginal comparisons (violet) against the reference posterior across training samples. Error bars show standard error of the mean over five MNPE training repetitions. Grey open circle: standard NPE baseline on dequantized discrete parameters at $N = 10^5$.
    \textbf{(c)}~Joint posterior: MNPE (green) versus reference posterior (blue). Black markers indicate ground-truth values.
    \textbf{(d)}~Calibration diagnostics at $10^5$ simulations. \emph{Left:}~SBC rank eCDF for the continuous parameters ($\gamma$, $\mu_1$, $\mu_2$), with 95\% uniform confidence band in grey. \emph{Center:}~ECE reliability diagram for the discrete parameters ($c_1$, $c_2$). \emph{Right:}~Expected calibration error for continuous (EoD, violet) and discrete (ECE, orange) parameters, averaged across five seeds and all marginals across budgets. Dashed lines indicate expected optimal calibration error for the given training budget.
    }
    \label{fig:queue}
\end{figure}

We apply MNPE to a tandem queueing simulator from operations research \cite{jackson1957networks}, consisting of two M/M/$c$ queues in series~\citep{gross_2008} where customers arrive at station~1 and, after being served, proceed to station~2. 
The inference problem has five parameters: three continuous arrival and service rates $\vtheta_c = (\gamma, \mu_1, \mu_2)$ and two discrete server counts $\vtheta_d = (c_1, c_2) \in \{2, \ldots, 6\}^2$ ($|\mathcal{D}| = 25$).
The simulator produces a five-dimensional observation comprising arrival counts, completion counts, and queue lengths over a fixed time horizon (see Appendix~\ref{app:exp_details} for details). 

For this simulator, no analytical posterior is available, but the closed-form likelihood allows us to construct a reference solution. Since gradient-based samplers such as NUTS~\citep{hoffman_gelman_2014} cannot directly handle discrete parameters, the 25 server-count configurations have to be marginalized out, and the resulting continuous posterior is multimodal, so NUTS does not converge on it. Conditional on the arrival rate, however, the two stations are independent, which reduces each configuration's evidence to one-dimensional integrals. We therefore compute the reference deterministically by quadrature and rejection sampling, and validate it against a sequential Monte Carlo reference (details in Appendix~\ref{app:exp_details}). Splitting the reference against itself yields a finite-sample C2ST floor of $0.502 \pm 0.003$, the level at which two samples from the same distribution are indistinguishable.
As an additional baseline for MNPE, we also train standard NPE on uniformly dequantized discrete parameters, recovering discrete values by flooring (Appendix~\ref{app:dequant}). 

We select the MNPE inference network architecture via hyperparameter search using Optuna \citep{akiba_optuna_2019}, optimizing the negative log-probability of the current posterior estimate on a held-out validation set \citep{lueckmann2021benchmarking}. The resulting network consists of a categorical MADE for the discrete parameters and a neural spline flow (NSF) for the continuous parameters (architecture details in Appendix~\ref{app:exp_details}). We train MNPE with up to $N=100{,}000$ simulations.

In the posterior predictive space, the MNPE-simulated observations match the reference posterior and concentrate around the true observation for a training budget of $N=100{,}000$ simulations (Fig.~\ref{fig:queue}a). 
Regarding posterior accuracy, we observe that for low simulation budgets, MNPE posteriors deviate substantially from the reference (measured by C2ST accuracy, Fig.~\ref{fig:queue}b), but the C2ST decreases monotonically with increasing simulation budget, reaching $0.536 \pm 0.006$ at $N = 100{,}000$ (mean $\pm$ s.e.m. over five training runs).
This is comparable to the MCMC-based references, which sit at $0.52$--$0.53$ against the same reference posterior, and clearly better than the dequantized NPE baseline at $0.630 \pm 0.008$ (Fig.~\ref{fig:queue}b). 
Increasing the budget to $1{,}000{,}000$ simulations further reduces the C2ST to $0.523$, matching the MCMC references (not shown). 

For a single example observation $\vx_o$ of arrival counts and queue lengths, MNPE recovers a posterior that closely matches the reference posterior across both continuous and discrete parameters and identifies the true server configuration $(c_1 = 2, c_2 = 3)$ as the dominant mode (Fig.~\ref{fig:queue}c). 
The MNPE posterior based on $N=100{,}000$ training simulations is well-calibrated: rank statistics for the continuous parameters are approximately uniform (Fig.~\ref{fig:queue}d, left), and reliability diagrams confirm that predicted class probabilities closely track empirical accuracy for the discrete parameters (Fig.~\ref{fig:queue}d, center). 
Across increasing MNPE training budgets, calibration improves consistently for the continuous dimensions, while it stays stable for the discrete dimensions, closely matching the expected optimal calibration (Fig.~\ref{fig:queue}d, right). 

In terms of timing, drawing $10{,}000$ samples from the MNPE posterior for a new observation takes $0.09$\,s, compared to $3.4$\,s for the SMC sampler and $42$--$90$\,s for per-configuration NUTS (Table~\ref{tab:runtime}). 
With a one-off simulation and training cost of about $500$\,s at $N = 100{,}000$, amortization pays off within $6$ to $150$ observations. Unlike likelihood-based alternatives, MNPE applies when only simulations are available.

\subsection{Hodgkin--Huxley simulator}

\begin{figure}[t]
    \centering
    \includegraphics[width=\textwidth]{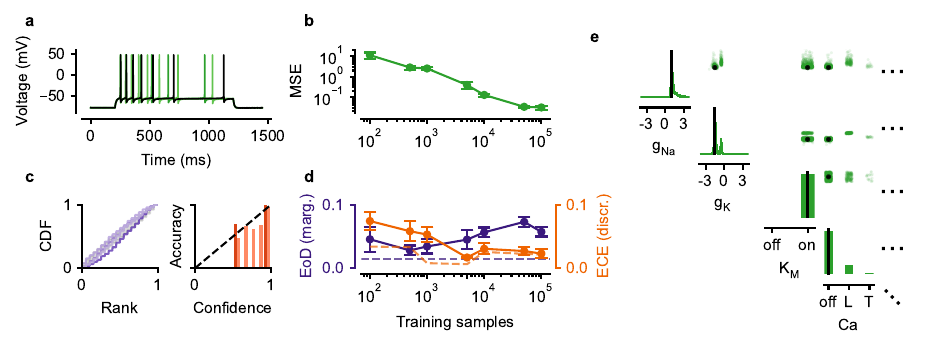}
    \caption{\textbf{MNPE on a Hodgkin-Huxley simulator.}
    \textbf{(a)} One observation (black) and two posterior predictive voltage traces (green) for a posterior trained on $100{,}000$ samples.
    \textbf{(b)} MSE (mode $\pm$std.) on 1k test samples. 
    \textbf{(c)} Simulation-based calibration for a posterior trained on $100{,}000$ samples,  
        \emph{Left:} for continuous parameters,
        \emph{Right:} for two discrete parameters. 
    \textbf{(d)} Calibration across training samples. Mean error over diagonal (EoD) on the continuous marginals and mean expected calibration error (ECE) on the discrete marginals. Dashed lines indicate the corresponding finite-sample baselines (Appendix~\ref{app:ece_baseline}). Error bars show std. across three random seeds.
    \textbf{(e)} One and two dimensional marginal posterior distribution for four selected parameters for the observation shown in a. Black dots/lines indicate ground truth parameters. Continuous parameters are presented on a normalized scale (see Fig.~\ref{app:fig:HH_posterior} for all dimensions).
    }
    \label{fig:hh}
\end{figure}

As a final example, we apply MNPE to the Hodgkin--Huxley simulator \citep{hodgkin1952quantitative}, which describes neural activity through the dynamics of ion channels. Different channels can produce qualitatively distinct behaviours, such as spike-frequency adaptation in the presence of $M$-type potassium channels ($K_m$) or diverse bursting patterns induced by calcium channels \citep{pospischil2008minimal}. Typically, inference focuses on the maximal conductances $g_i$ of the channels, and channel-identities are assumed to be known. However, in practice, the exact combination of channels in a neuron is often unknown. Here, we apply MNPE to simultaneously infer continuous channel densities and categorical channel identities. While a large variety of ion channels exists \citep{podlaski2017mapping}, we focused on a minimal set consisting of $Na$ and $K$ channels, optional $K_m$, and $L$- and $T$-type calcium channels ($Ca_L$, $Ca_T$), which together yield rich dynamical regimes \citep{pospischil2008minimal}.

We implemented the simulator using the \textsc{Jaxley} simulator \citep{deistler2025jaxley} and added observational noise to the simulated voltage trace (Fig.~\ref{fig:hh}a). To reduce the dimensionality of the time-series we summarized it by a set of 14 features, such as first and second moments computed over different time intervals, mirroring previous work \citep{goncalves2020training} (details in Appendix~\ref{app:exp_details}). This results in an inference problem with four continuous parameters and two discrete variables: The continuous variables modulate the sodium, potassium, and leak conductances ($g_{Na}, g_K, g_L$), as well as the leak reversal potential ($E_L$). The two discrete parameters indicate the presence of slow potassium dynamics $K_m$ and select the calcium channel type ($Ca_L$, $Ca_T$, or none) with fixed parametrization.

The Hodgkin--Huxley likelihood is intractable, so unlike the previous two examples this one has no reference posterior.
We therefore assess the posterior by its calibration and by its predictive adequacy, both of which require only simulations.
We train MNPE on $100{,}000$ simulations and perform inference given a synthetically generated observation. We found that samples from the posterior closely match the observation qualitatively, with similar spiking characteristics (Fig.~\ref{fig:hh}a). 
We then evaluate the performance of MNPE with respect to the amount of available training data. As expected, the mean squared error (MSE) of the summary statistics decreases with increasing numbers of training samples (Fig.~\ref{fig:hh}b).

Next, we evaluate the calibration of the MNPE posterior on $500$ held-out prior draws.
Calibration of the discrete parameters improves with the training budget, from an expected calibration error of $0.074$ at $10^2$ simulations to $0.023$ at $10^5$, where it approaches its finite-sample baseline (\Figref{fig:hh}d).
For the continuous parameters the rank statistics stay slightly above their baseline at every budget (\Figref{fig:hh}c left, d).
The rank CDFs are bowed rather than U-shaped, and in opposite directions for different parameters, which indicates a small residual bias of the continuous marginals rather than over- or underconfidence.
The posterior predictive error decreases over the same range (\Figref{fig:hh}b), so the posterior becomes more informative while this bias persists. 
Interestingly, the discrete parameter for $K_m$ mainly occupies two bins in the ECE diagram, which indicates that the posterior has either high or low confidence, but is well calibrated in both cases (Fig.~\ref{fig:hh}c, right).
Having a joint posterior over continuous and discrete parameters in this situation allows to draw conclusions about the presence of specific ion channels and additionally enables the analysis of compensatory mechanisms and interactions between ion channels. For example, for this particular simulation of  a low-frequency spiking neuron with temporal adaptation, we observe that the presence of an $L$-type calcium channel may be associated with higher potassium conductance $g_K$ (Fig.~\ref{fig:hh}e).

\section{Discussion}
\label{sec:discussion}

We presented MNPE, an extension of neural posterior estimation to simulators with mixed discrete and continuous parameters. 
Across three examples of increasing complexity, the approach produces posterior estimates that match reference posteriors where available and remain well-calibrated for both continuous and discrete parameter dimensions. 
To assess calibration in the mixed setting, we combined rank-based diagnostics for continuous parameters with classification-based metrics for discrete parameters and derived finite-sample baselines that allow reliable interpretation of both.

\paragraph{Related work.}
Our work is inspired by previous work on performing inference in simulators with mixed simulation \emph{outputs}, namely mixed likelihood estimation \citep[MNLE,][]{boelts_mnle_2022}. While MNLE factorizes the observations (i.e., \emph{likelihoods}) into discrete and continuous components, MNPE targets applications with mixed \emph{parameter spaces}, and directly the posterior. In addition, our work extends the inference network to multiple categorical variables by using categorical MADEs.
In a similar spirit to our discrete--continuous factorization, dedicated methods have been proposed for hierarchical models. \cite{rodrigues2021hnpe} introduce an NPE method for hierarchical model structures in which observations share certain global parameters, while \cite{habermann2025amortized} extend this to general multilevel hierarchical models. Although both approaches share the core idea of factorization with separate networks for global and local parameters, they are restricted to continuous parameters. Hybrid discrete--continuous generative models have also been developed outside SBI. For example, TabDDPM combines Gaussian and multinomial diffusion processes, CoDi couples separate continuous and discrete diffusion models, and DLT and TabDiff define joint mixed-type diffusion processes \citep{kotelnikov2023tabddpm,lee2023codi,levi2023dlt, shi2025tabdiff}. These methods primarily target data synthesis, whereas MNPE targets amortized posterior inference for simulators.

Several recent methods address the mixed inference problem using diffusion models. \cite{ghiglino2026diffusion_mixed_estimation} adopt the same conditional factorization
as MNPE but model both components with diffusion processes, requiring a Riemannian continuous embedding for the discrete space.
\cite{schroeder2024simultaneous} treat $\vtheta_d$ as a model index and use simple mixture-based posteriors to perform inference over both, alternative model components and their associated parameters.
\cite{gloeckler2026scalable} extend joint inference setting to a transformer-based encoder-decoder architecture combined with a diffusion process over the continuous parameters, resulting in a more flexible but substantially heavier framework.
MNPE's contribution is a lightweight and modular NPE construction that represents categorical parameters directly with an autoregressive classifier and combines it with a standard conditional estimator for the continuous parameters. Unlike diffusion-based alternatives, the flow-based implementation used here provides tractable joint log-probabilities and ancestral sampling without iterative denoising.
Importantly, MNPE's discrete--continuous factorization is agnostic to the choice of underlying generative model for each component. Any discrete or continuous estimator, including diffusion-based ones, can in principle be substituted within our framework, making MNPE a general compositional scheme rather than one tied to a specific backbone.
However, for high-dimensional posteriors, large simulation budgets, or scenarios where inference time is less critical, diffusion-based methods may still be advantageous.
In the direction of foundation models, in-context learning approaches \citep{vetter2025effortless} show promise for heterogeneous feature types and could motivate future alternatives.

While MCMC methods are a natural baseline for posterior inference in the setting of available likelihoods, handling mixed discrete and continuous parameters remains challenging. Discontinuous HMC~\citep{nishimura2020discontinuousHMC} embeds discrete parameters into a continuous space and simulates Hamiltonian dynamics on a piecewise-smooth density, but is limited to ordinal discrete parameters and does not generalize to arbitrary discrete state spaces. Mixed HMC~\citep{zhou2020mixed} evolves discrete and continuous variables jointly within HMC trajectories, yet still requires a differentiable likelihood and offers no advantage over analytically marginalizing out the discrete dimensions. Furthermore, marginalization cost scales as $O(|\mathcal{D}|)$ per MCMC step, which becomes increasingly costly for larger discrete spaces, a limitation that MNPE does not share. 
Explicit enumeration is feasible only for small discrete state spaces and becomes combinatorially costly as dimensionality grows. MNPE avoids enumeration by modeling the discrete posterior autoregressively and sharing one continuous estimator across states. Its benefits are therefore greatest for larger discrete spaces, provided sufficient coverage during training (Appendix~\ref{app:scaling}).
Both MCMC approaches furthermore require tractable likelihood and gradient evaluations and do not amortize---every new observation demands a full sampling run. MNPE, in comparison, can perform inference for any black-box simulator and directly draws samples from the posterior distribution in a single forward pass, thereby amortizing the cost of inference.

\paragraph{Limitations.}
MNPE does not directly support transdimensional posteriors defined over a union of parameter spaces with different dimensions, as arise when the number of objects or components is unknown \citep{green1995reversible,davies2025transdimensional}. Such problems could be embedded into a fixed-dimensional representation by introducing a maximum number of objects and discrete presence indicators. However, this introduces inactive parameters and permutation symmetries, and requires additional treatment of masking, exchangeability, and label switching. Variable-cardinality inference, therefore, lies outside of the scope of the presented method.
As our work is a method in the NPE family, it also inherits its weaknesses: it may need a large training dataset to provide accurate estimations for complex posterior structures (e.g. \citep{deistler2022energy}), it is susceptible to model misspecification \citep{cannon2022investigating,kelly2025misspecification}, and while it targets the posterior across the full prior range, one might only be interested in a specific observation. In such a case, multi-round NPE approaches \citep{greenberg2019automatic} can be adopted.
The calibration checks reported in the main text assess each marginal separately.
The full mixed posterior is not appropriate for standard expected-coverage checks \citep{deistler2022truncated}, but the coupling between the discrete and continuous factors can be tested directly (details in Appendix~\ref{app:conditional}).
Furthermore, the empirical ECE baseline relies on the half-normal approximation to the binomial, which can underestimate the noise floor at very small test set sizes (Appendix~\ref{app:ece_baseline}).

\paragraph{Conclusion.}
Simulators with discrete parameters are prevalent throughout the sciences and engineering; yet their specifics are often overlooked when designing inference and calibration methods for simulation-based inference. We presented a strategy to build inference networks addresses this. We showed the effectiveness of this method across multiple examples and we will provide an open source implementation upon acceptance.

\section*{Acknowledgements}
We thank all members of the Mackelab for discussions and feedback on the manuscript. 
This work was funded by the German Research Foundation (DFG) under Germany’s Excellence Strategy – EXC number 2064/1 – 390727645 and SFB 1233 `Robust Vision' (276693517), the European Union (ERC, DeepCoMechTome, 101089288), and the Boehringer Ingelheim AI \& Data Science Fellowship Program. JaB is supported by the appliedAI Institute for Europe. JoB is a member of the International Max Planck Research School for Intelligent Systems (IMPRS-IS).

\bibliography{references}

@article{boelts_mnle_2022,
  title={Flexible and efficient simulation-based inference for models of decision-making},
  author={Boelts, Jan and Lueckmann, Jan-Matthis and Gao, Richard and Macke, Jakob H},
  journal={eLife},
  volume={11},
  pages={e77220},
  year={2022},
  publisher={eLife Sciences Publications, Ltd},
}

@article{jarrett_1979,
  title={A note on the intervals between coal-mining disasters},
  author={Jarrett, R. G.},
  journal={Biometrika},
  volume={66},
  number={1},
  pages={191--193},
  year={1979},
  publisher={Oxford University Press},
}

@article{raftery_akman_1986,
  title={Bayesian analysis of a {P}oisson process with a change-point},
  author={Raftery, Adrian E. and Akman, Varol E.},
  journal={Biometrika},
  volume={73},
  number={1},
  pages={85--89},
  year={1986},
  publisher={Oxford University Press},
}

@article{pymc_2023,
  title={{PyMC}: a modern, and comprehensive probabilistic programming framework in {P}ython},
  author={Abril-Pla, Oriol and Andreani, Virgile and Carroll, Colin and Dong, Larry and Fonnesbeck, Christopher J. and Kochurov, Maxim and Kumar, Ravin and Lao, Junpeng and Luhmann, Christian C. and Martin, Osvaldo A. and Osthege, Michael and Vieira, Ricardo and Wiecki, Thomas and Zinkov, Robert},
  journal={PeerJ Computer Science},
  volume={9},
  pages={e1516},
  year={2023},
  publisher={PeerJ Inc.},
}

@article{hoffman_gelman_2014,
  title={The {N}o-{U}-{T}urn sampler: adaptively setting path lengths in {H}amiltonian {M}onte {C}arlo},
  author={Hoffman, Matthew D. and Gelman, Andrew},
  journal={Journal of Machine Learning Research},
  volume={15},
  pages={1593--1623},
  year={2014},
}

@article{papamakarios2016fast,
  title={Fast $\varepsilon$-free inference of simulation models with {B}ayesian conditional density estimation},
  author={Papamakarios, George and Murray, Iain},
  journal={Advances in Neural Information Processing Systems},
  volume={29},
  year={2016},
}

@inproceedings{greenberg2019automatic,
  title={Automatic posterior transformation for likelihood-free inference},
  author={Greenberg, David and Nonnenmacher, Marcel and Macke, Jakob},
  booktitle={International Conference on Machine Learning},
  pages={2404--2414},
  year={2019},
  organization={PMLR},
}

@inproceedings{lueckmann2021benchmarking,
  title={Benchmarking simulation-based inference},
  author={Lueckmann, Jan-Matthis and Boelts, Jan and Greenberg, David and Goncalves, Pedro and Macke, Jakob},
  booktitle={International Conference on Artificial Intelligence and Statistics},
  pages={343--351},
  year={2021},
  organization={PMLR},
}

@article{goncalves2020training,
  title={Training deep neural density estimators to identify mechanistic models of neural dynamics},
  author={Gon{\c{c}}alves, Pedro J and Lueckmann, Jan-Matthis and Deistler, Michael and Nonnenmacher, Marcel and {\"O}cal, Kaan and Bassetto, Giacomo and Chintaluri, Chaitanya and Podlaski, William F and Haddad, Sara A and Vogels, Tim P and others},
  journal={eLife},
  volume={9},
  pages={e56261},
  year={2020},
  publisher={eLife Sciences Publications Limited},
}

@article{cranmer2020frontier,
  title={The frontier of simulation-based inference},
  author={Cranmer, Kyle and Brehmer, Johann and Louppe, Gilles},
  journal={Proceedings of the National Academy of Sciences},
  volume={117},
  number={48},
  pages={30055--30062},
  year={2020},
  publisher={National Academy of Sciences},
}

@article{talts2018validating,
  title={Validating {B}ayesian inference algorithms with simulation-based calibration},
  author={Talts, Sean and Betancourt, Michael and Simpson, Daniel and Vehtari, Aki and Gelman, Andrew},
  journal={arXiv preprint arXiv:1804.06788},
  year={2018},
}

@article{deistler2022energy,
  title={Energy-efficient network activity from disparate circuit parameters},
  author={Deistler, Michael and Macke, Jakob H and Gon{\c{c}}alves, Pedro J},
  journal={Proceedings of the National Academy of Sciences},
  volume={119},
  number={44},
  pages={e2207632119},
  year={2022},
  publisher={National Academy of Sciences},
}

@article{cannon2022investigating,
  title={Investigating the impact of model misspecification in neural simulation-based inference},
  author={Cannon, Patrick and Ward, Daniel and Schmon, Sebastian M},
  journal={arXiv preprint arXiv:2209.01845},
  year={2022},
}

@inproceedings{deistler2022truncated,
  title={Truncated proposals for scalable and hassle-free simulation-based inference},
  author={Deistler, Michael and Goncalves, Pedro J. and Macke, Jakob H.},
  booktitle={Advances in Neural Information Processing Systems},
  year={2022},
}

@article{radev2020bayesflow,
  title={{BayesFlow}: Learning complex stochastic models with invertible neural networks},
  author={Radev, Stefan T and Mertens, Ulf K and Voss, Andreas and Ardizzone, Lynton and K{\"o}the, Ullrich},
  journal={IEEE Transactions on Neural Networks and Learning Systems},
  volume={33},
  number={4},
  pages={1452--1466},
  year={2020},
  publisher={IEEE},
}

@inproceedings{simons2023neural,
  title={Neural score estimation: Likelihood-free inference with conditional score based diffusion models},
  author={Simons, Jack and Sharrock, Louis and Liu, Song and Beaumont, Mark},
  booktitle={Symposium on Advances in Approximate Bayesian Inference},
  year={2023},
}

@InProceedings{germain2015made,
  title={{MADE}: Masked autoencoder for distribution estimation},
  author={Germain, Mathieu and Gregor, Karol and Murray, Iain and Larochelle, Hugo},
  booktitle={International Conference on Machine Learning},
  pages={881--889},
  year={2015},
  volume={37},
  series={Proceedings of Machine Learning Research},
  publisher={PMLR},
}

@article{dax2021real,
  title={Real-time gravitational wave science with neural posterior estimation},
  author={Dax, Maximilian and Green, Stephen R and Gair, Jonathan and Macke, Jakob H and Buonanno, Alessandra and Sch{\"o}lkopf, Bernhard},
  journal={Physical Review Letters},
  volume={127},
  number={24},
  pages={241103},
  year={2021},
  publisher={APS},
}

@article{deistler2025simulation,
  title={Simulation-based inference: {A} practical guide},
  author={Deistler, Michael and Boelts, Jan and Steinbach, Peter and Moss, Guy and Moreau, Thomas and Gloeckler, Manuel and Rodrigues, Pedro L. C. and Linhart, Julia and Lappalainen, Janne K. and Miller, Benjamin Kurt and others},
  journal={arXiv preprint arXiv:2508.12939},
  year={2025},
}

@article{deistler2025jaxley,
  title={Jaxley: differentiable simulation enables large-scale training of detailed biophysical models of neural dynamics},
  author={Deistler, Michael and Kadhim, Kyra L and Pals, Matthijs and Beck, Jonas and Huang, Ziwei and Gloeckler, Manuel and Lappalainen, Janne K and Schr{\"o}der, Cornelius and Berens, Philipp and Gon{\c{c}}alves, Pedro J and others},
  journal={Nature Methods},
  volume={22},
  pages={2649--2657},
  year={2025},
  publisher={Nature Publishing Group},
}

@inproceedings{guo2017calibration,
  title={On calibration of modern neural networks},
  author={Guo, Chuan and Pleiss, Geoff and Sun, Yu and Weinberger, Kilian Q},
  booktitle={International Conference on Machine Learning},
  pages={1321--1330},
  year={2017},
  organization={PMLR},
}

@article{pospischil2008minimal,
  title={Minimal {H}odgkin--{H}uxley type models for different classes of cortical and thalamic neurons},
  author={Pospischil, Martin and Toledo-Rodriguez, Maria and Monier, Cyril and Piwkowska, Zuzanna and Bal, Thierry and Fr{\'e}gnac, Yves and Markram, Henry and Destexhe, Alain},
  journal={Biological Cybernetics},
  volume={99},
  number={4},
  pages={427--441},
  year={2008},
  publisher={Springer},
}

@article{boeltsdeistler2025reloaded,
  title={sbi reloaded: {A} toolkit for simulation-based inference workflows},
  author={Boelts, Jan and Deistler, Michael and Gloeckler, Manuel and {Tejero-Cantero}, {\'A}lvaro and Lueckmann, Jan-Matthis and Moss, Guy and Steinbach, Peter and Moreau, Thomas and Muratore, Fabio and Linhart, Julia and Durkan, Conor and Vetter, Julius and Miller, Benjamin Kurt and Herold, Maternus and Ziaeemehr, Abolfazl and Pals, Matthijs and Gruner, Theo and Bischoff, Sebastian and Krouglova, Nastya and Gao, Richard and Lappalainen, Janne K. and Mucs{\'a}nyi, B{\'a}lint and Pei, Felix and Schulz, Auguste and Stefanidi, Zinovia and Rodrigues, Pedro and Schr{\"o}der, Cornelius and Zaid, Faried Abu and Beck, Jonas and Kapoor, Jaivardhan and Greenberg, David S. and Gon{\c c}alves, Pedro J. and Macke, Jakob H.},
  year={2025},
  journal={Journal of Open Source Software},
  volume={10},
  number={108},
  pages={7754},
}

@inproceedings{akiba_optuna_2019,
  title={Optuna: {A} next-generation hyperparameter optimization framework},
  author={Akiba, Takuya and Sano, Shotaro and Yanase, Toshihiko and Ohta, Takeru and Koyama, Masanori},
  booktitle={Proceedings of the 25th ACM SIGKDD International Conference on Knowledge Discovery \& Data Mining},
  pages={2623--2631},
  year={2019},
  publisher={Association for Computing Machinery},
}

@book{gross_2008,
  author={Gross, Donald and Shortle, John F. and Thompson, James M. and Harris, Carl M.},
  title={Fundamentals of Queueing Theory},
  edition={4th},
  publisher={Wiley-Interscience},
  year={2008},
}

@article{phan2019_numpyro,
  title={Composable effects for flexible and accelerated probabilistic programming in {NumPyro}},
  author={Phan, Du and Pradhan, Neeraj and Jankowiak, Martin},
  journal={arXiv preprint arXiv:1912.11554},
  year={2019},
}

@article{gneiting2007probabilistic,
  title={Probabilistic forecasts, calibration and sharpness},
  author={Gneiting, Tilmann and Balabdaoui, Fadoua and Raftery, Adrian E},
  journal={Journal of the Royal Statistical Society Series B: Statistical Methodology},
  volume={69},
  number={2},
  pages={243--268},
  year={2007},
  publisher={Oxford University Press},
}

@article{cook2006validation,
  title={Validation of software for {B}ayesian models using posterior quantiles},
  author={Cook, Samantha R and Gelman, Andrew and Rubin, Donald B},
  journal={Journal of Computational and Graphical Statistics},
  volume={15},
  number={3},
  pages={675--692},
  year={2006},
  publisher={Taylor \& Francis},
}

@article{hodgkin1952quantitative,
  title={A quantitative description of membrane current and its application to conduction and excitation in nerve},
  author={Hodgkin, Alan L and Huxley, Andrew F},
  journal={The Journal of Physiology},
  volume={117},
  number={4},
  pages={500--544},
  year={1952},
}

@article{podlaski2017mapping,
  title={Mapping the function of neuronal ion channels in model and experiment},
  author={Podlaski, William F and Seeholzer, Alexander and Groschner, Lukas N and Miesenb{\"o}ck, Gero and Ranjan, Rajnish and Vogels, Tim P},
  journal={eLife},
  volume={6},
  pages={e22152},
  year={2017},
  publisher={eLife Sciences Publications, Ltd},
}

@misc{allen_database,
  author={{Allen Institute for Brain Science}},
  title={Allen cell types database},
  howpublished={\url{http://celltypes.brain-map.org}},
  year={2016},
}

@inproceedings{schroeder2024simultaneous,
  title={Simultaneous identification of models and parameters of scientific simulators},
  author={Schr{\"o}der, Cornelius and Macke, Jakob H},
  booktitle={International Conference on Machine Learning},
  pages={43895--43927},
  year={2024},
  organization={PMLR},
}

@article{zhou2020mixed,
  title={Mixed {H}amiltonian {M}onte {C}arlo for mixed discrete and continuous variables},
  author={Zhou, Guangyao},
  journal={Advances in Neural Information Processing Systems},
  volume={33},
  pages={17094--17104},
  year={2020},
}

@techreport{friedman2004multivariate,
  title={On multivariate goodness-of-fit and two-sample testing},
  author={Friedman, Jerome},
  year={2004},
  institution={Stanford Linear Accelerator Center},
}

@inproceedings{lopez-paz2017revisiting,
  title={Revisiting classifier two-sample tests},
  author={Lopez-Paz, David and Oquab, Maxime},
  booktitle={International Conference on Learning Representations},
  year={2017},
}

@article{gloeckler2026scalable,
  title={Scalable simulation-based model inference with test-time complexity control},
  author={Gloeckler, Manuel and Manzano-Patr{\'o}n, J. P. and Sotiropoulos, Stamatios N. and Schr{\"o}der, Cornelius and Macke, Jakob H.},
  journal={arXiv preprint arXiv:2603.15292},
  year={2026},
}

@article{habermann2025amortized,
  title={Amortized {B}ayesian multilevel models},
  author={Habermann, Daniel and Schmitt, Marvin and K{\"u}hmichel, Lars and Bulling, Andreas and Radev, Stefan T. and B{\"u}rkner, Paul-Christian},
  journal={Bayesian Analysis},
  pages={1--30},
  year={2025},
  note={Advance publication},
}

@article{rodrigues2021hnpe,
  title={{HNPE}: Leveraging global parameters for neural posterior estimation},
  author={Rodrigues, Pedro and Moreau, Thomas and Louppe, Gilles and Gramfort, Alexandre},
  journal={Advances in Neural Information Processing Systems},
  volume={34},
  pages={13432--13443},
  year={2021},
}

@article{ghiglino2026diffusion_mixed_estimation,
  title={Do diffusion models dream of electric planes? {D}iscrete and continuous simulation-based inference for aircraft design},
  author={Ghiglino, Aurelien and Elenius, Daniel and Roy, Anirban and Kaur, Ramneet and Acharya, Manoj and Samplawski, Colin and Matejek, Brian and Jha, Susmit and Alonso, Juan and Cobb, Adam},
  journal={arXiv preprint arXiv:2603.13284},
  year={2026},
}

@article{nishimura2020discontinuousHMC,
  title={Discontinuous {H}amiltonian {M}onte {C}arlo for discrete parameters and discontinuous likelihoods},
  author={Nishimura, Akihiko and Dunson, David B and Lu, Jianfeng},
  journal={Biometrika},
  volume={107},
  number={2},
  pages={365--380},
  year={2020},
  publisher={Oxford University Press},
}

@article{kelly2025misspecification,
  title={Simulation-based {B}ayesian inference under model misspecification},
  author={Kelly, Ryan P. and Warne, David J. and Frazier, David T. and Nott, David J. and Gutmann, Michael U. and Drovandi, Christopher},
  journal={arXiv preprint arXiv:2503.12315},
  year={2025},
}

@inproceedings{shi2025tabdiff,
  title={{TabDiff}: A mixed-type diffusion model for tabular data generation},
  author={Shi, Juntong and Xu, Minkai and Hua, Harper and Zhang, Hengrui and Ermon, Stefano and Leskovec, Jure},
  booktitle={International Conference on Learning Representations},
  year={2025},
}

@inproceedings{vetter2025effortless,
  title={Effortless, simulation-efficient {B}ayesian inference using tabular foundation models},
  author={Vetter, Julius and Gloeckler, Manuel and Gedon, Daniel and Macke, Jakob H.},
  booktitle={Advances in Neural Information Processing Systems},
  year={2025},
}

@article{fengler2021LAN,
  title={Likelihood approximation networks ({LANs}) for fast inference of simulation models in cognitive neuroscience},
  author={Fengler, Alexander and Govindarajan, Lakshmi N and Chen, Tony and Frank, Michael J},
  journal={eLife},
  volume={10},
  pages={e65074},
  year={2021},
}

@article{mckinley2014simulation,
  title={Simulation-based {B}ayesian inference for epidemic models},
  author={McKinley, Trevelyan J and Ross, Joshua V and Deardon, Rob and Cook, Alex R},
  journal={Computational Statistics \& Data Analysis},
  volume={71},
  pages={434--447},
  year={2014},
  publisher={Elsevier},
}

@inproceedings{altamirano2023robust,
  title={Robust and scalable {B}ayesian online changepoint detection},
  author={Altamirano, Matias and Briol, Francois-Xavier and Knoblauch, Jeremias},
  booktitle={ICML Workshop on Structured Probabilistic Inference \& Generative Modeling},
  year={2023},
}

@article{adams2007bayesian,
  title={{B}ayesian online changepoint detection},
  author={Adams, Ryan Prescott and MacKay, David J. C.},
  journal={arXiv preprint arXiv:0710.3742},
  year={2007},
}

@inproceedings{linderman17a,
  title={{B}ayesian learning and inference in recurrent switching linear dynamical systems},
  author={Linderman, Scott and Johnson, Matthew and Miller, Andrew and Adams, Ryan and Blei, David and Paninski, Liam},
  booktitle={International Conference on Artificial Intelligence and Statistics},
  pages={914--922},
  year={2017},
  volume={54},
  series={Proceedings of Machine Learning Research},
}

@article{fu2024simultaneous,
  title={Simultaneous identification of changepoints and model parameters in switching dynamical systems},
  author={Fu, Xiaoming and Fan, Kai and Zozmann, Heinrich and Sch{\"u}ler, Lennart and Calabrese, Justin M},
  journal={bioRxiv},
  pages={2024.01.30.577909},
  year={2024},
}

@article{radev2021amortized,
  title={Amortized {B}ayesian model comparison with evidential deep learning},
  author={Radev, Stefan T and D'Alessandro, Marco and Mertens, Ulf K and Voss, Andreas and Koethe, Ullrich and Buerkner, Paul-Christian},
  journal={IEEE Transactions on Neural Networks and Learning Systems},
  volume={34},
  number={8},
  pages={4903--4917},
  year={2021},
  publisher={IEEE},
}

@article{jackson1957networks,
  title={Networks of waiting lines},
  author={Jackson, James R},
  journal={Operations Research},
  volume={5},
  number={4},
  pages={518--521},
  year={1957},
  publisher={INFORMS},
}

@article{von2022mental,
  title={Mental speed is high until age 60 as revealed by analysis of over a million participants},
  author={von Krause, Mischa and Radev, Stefan T and Voss, Andreas},
  journal={Nature Human Behaviour},
  volume={6},
  number={5},
  pages={700--708},
  year={2022},
  publisher={Nature Publishing Group},
}

@article{kuhmichel2026bayesflow,
  title={{BayesFlow} 2.0: Multi-backend amortized {B}ayesian inference in {P}ython},
  author={K{\"u}hmichel, Lars and Huang, Jerry M and Pratz, Valentin and Arruda, Jonas and Olischl{\"a}ger, Hans and Habermann, Daniel and Kucharsky, Simon and Elsem{\"u}ller, Lasse and Mishra, Aayush and Bracher, Niels and others},
  journal={arXiv preprint arXiv:2602.07098},
  year={2026},
}

@article{kumar2019verified,
  title={Verified uncertainty calibration},
  author={Kumar, Ananya and Liang, Percy S and Ma, Tengyu},
  journal={Advances in Neural Information Processing Systems},
  volume={32},
  year={2019},
}

@inproceedings{roelofs2022mitigating,
  title={Mitigating bias in calibration error estimation},
  author={Roelofs, Rebecca and Cain, Nicholas and Shlens, Jonathon and Mozer, Michael C},
  booktitle={International Conference on Artificial Intelligence and Statistics},
  pages={4036--4054},
  year={2022},
  organization={PMLR},
}

@article{widmann2019calibration,
  title={Calibration tests in multi-class classification: {A} unifying framework},
  author={Widmann, David and Lindsten, Fredrik and Zachariah, Dave},
  journal={Advances in Neural Information Processing Systems},
  volume={32},
  year={2019},
}

@inproceedings{vaicenavicius2019evaluating,
  title={Evaluating model calibration in classification},
  author={Vaicenavicius, Juozas and Widmann, David and Andersson, Carl and Lindsten, Fredrik and Roll, Jacob and Sch{\"o}n, Thomas},
  booktitle={International Conference on Artificial Intelligence and Statistics},
  pages={3459--3467},
  year={2019},
  organization={PMLR},
}

@inproceedings{nixon2019measuring,
  title={Measuring calibration in deep learning},
  author={Nixon, Jeremy and Dusenberry, Michael W and Zhang, Linchuan and Jerfel, Ghassen and Tran, Dustin},
  booktitle={CVPR Workshops},
  year={2019},
}

@inproceedings{kotelnikov2023tabddpm,
  title     = {TabDDPM: Modelling Tabular Data with Diffusion Models},
  author    = {Kotelnikov, Akim and Baranchuk, Dmitry and
               Rubachev, Ivan and Babenko, Artem},
  booktitle = {Proceedings of the 40th International Conference
               on Machine Learning},
  series    = {Proceedings of Machine Learning Research},
  volume    = {202},
  pages     = {17564--17579},
  year      = {2023}
}

@inproceedings{lee2023codi,
  title     = {CoDi: Co-evolving Contrastive Diffusion Models
               for Mixed-type Tabular Synthesis},
  author    = {Lee, Chaejeong and Kim, Jayoung and Park, Noseong},
  booktitle = {Proceedings of the 40th International Conference
               on Machine Learning},
  series    = {Proceedings of Machine Learning Research},
  volume    = {202},
  pages     = {18940--18956},
  year      = {2023}
}

@inproceedings{levi2023dlt,
  title={DLT: Conditioned Layout Generation with Joint Discrete-Continuous Diffusion Layout Transformer},
  author={Levi, Elad and Brosh, Eli and Mykhailych, Mykola and Perez, Meir},
  booktitle={Proceedings of the IEEE/CVF International Conference on Computer Vision},
  pages={2106--2115},
  year={2023},
}

@article{green1995reversible,
  author  = {Green, Peter J.},
  title   = {Reversible Jump Markov Chain Monte Carlo Computation
             and Bayesian Model Determination},
  journal = {Biometrika},
  year    = {1995},
  volume  = {82},
  number  = {4},
  pages   = {711--732},
}

@inproceedings{davies2025transdimensional,
  author    = {Davies, Laurence and MacKinlay, Daniel and
               Oliveira, Rafael and Sisson, Scott A.},
  title     = {Variational Transdimensional Inference},
  booktitle = {Advances in Neural Information Processing Systems},
  volume    = {38},
  year      = {2025}
}
\bibliographystyle{tmlr}

\clearpage
\appendix

\setcounter{section}{0}
\renewcommand{\thesection}{\Alph{section}}

\renewcommand{\thefigure}{\Alph{section}-\arabic{figure}}
\setcounter{figure}{0}
\renewcommand{\thetable}{\Alph{section}-\arabic{table}}
\setcounter{table}{0}
\renewcommand{\theequation}{\Alph{section}-\arabic{equation}}
\setcounter{equation}{0} %
\counterwithin*{figure}{section}
\counterwithin*{table}{section}

\addcontentsline{toc}{section}{Appendix}
\setcounter{tocdepth}{2} %

\begin{center}
    \Large\bfseries Appendix Contents
\end{center}

\appendixtoccolor
\startcontents[appendix]
\printcontents[appendix]{}{1}{}
\restorelinkcolor
\newpage

\section{Calibration checks}

\subsection{Empirical baselines for discrete calibration checks}
\label{app:ece_baseline}

For the continuous parameters, the calibration baseline is straightforward: uniform SBC ranks define an estimator-independent ideal, and the expected EoD under perfect calibration follows analytically from the number of test cases $N$ and rank bins $B$ alone.
For the discrete parameters, the situation is different.
The $\ell_1$-ECE is a positively biased estimator on finite test sets \citep{kumar2019verified,roelofs2022mitigating}: per-bin accuracies fluctuate around the true confidence by binomial sampling noise, so even a perfectly calibrated posterior produces a strictly positive ECE.
The magnitude of this noise floor depends on how the test cases distribute across the confidence bins, which is determined by the posterior estimator itself --- unlike the EoD baseline, the ECE baseline is therefore inherently posterior-dependent.
This section derives the baseline we propose to use.

The check requires a choice of the probabilities it consumes.
The reliability diagrams and ECE of Section~\ref{subsec:calibration} are defined on the per-dimension marginals $q(\theta_{d_i} \mid \vx)$, but for multiple discrete dimensions the autoregressive head returns the conditional factors $q(\theta_{d_i} \mid \vtheta_{d_{<i}}, \vx)$, which must not be evaluated at the true preceding values, since those are unavailable at inference time. We therefore marginalize over the preceding dimensions by enumerating the prefix configurations where their number is small, as in the queueing model, and by ancestral Monte Carlo at the larger dimension counts of Appendix~\ref{app:scaling}.

To quantify the noise floor for the ECE, we consider $B$ equal-width confidence bins centered at $p_b$ and $n_b$ test cases per bin (approximating the mean confidence in the bin). 
Under perfect calibration, each test case in bin $b$ is classified correctly with probability $p_b$. The count of correct predictions therefore follows a binomial distribution $n_b \cdot \operatorname{acc}(b) \sim \operatorname{Binomial}(n_b,p_b)$ and the per-bin contribution to the expected calibration error is 
\begin{equation*}
    \frac{n_b}{N}\,\mathbb{E}\bigl[|\operatorname{acc}(b) - p_b|\bigr] =\frac{n_b}{N} \sum_{k=0}^{n_b} \bigl|\frac{k}{n_b} - p_b\bigr|\binom{n_b}{k} p_b^k (1-p_b)^{n_b-k}~.
\end{equation*}
Since the calculation of this sum can get expensive for large $n_b$, we approximate it via a half-normal distribution. 
A commonly used rule of thumb is $n_b p_b \geq 5$ and $n_b(1-p_b) \geq 5$ for a reliable approximation. 
Applying this approximation yields 
\begin{equation*}
    \mathbb{E}\bigl[|\operatorname{acc}(b) - p_b|\bigr]
    \;\approx\; \sqrt{\frac{2}{\pi}}\;\sqrt{\frac{p_b(1 - p_b)}{n_b}}\,,
\end{equation*}
and the expected ECE under perfect calibration reduces to
\begin{equation*}
    \mathbb{E}[\operatorname{ECE}]
    \;=\; \frac{1}{N}\,\sqrt{\frac{2}{\pi}}\;
    \sum_{b}\,\sqrt{\,n_b \cdot p_b(1 - p_b)}\,.
\end{equation*}
This formula depends on the bin occupancies $\{n_b\}$ under the generally unknown true posterior. 
Assuming a uniform occupancy produces a higher baseline, because the high-variance bins in the center of the interval are weighted equally with the low-variance ones. 
Instead, we suggest using a tighter baseline that uses empirical bin counts obtained from the approximate posterior estimate \citep{widmann2019calibration,vaicenavicius2019evaluating}. 
Importantly, the posterior estimator is used only for obtaining bin occupancy, not for the confidences themselves ($p_b$ is defined by the center of each bin). 
Note that because the bin counts depend on the posterior estimator, this baseline varies with the confidence of the posterior estimator, e.g., with training budgets.
While previous work replaced the equal-width bins with equal-mass (adaptive quantile) bins \citep{nixon2019measuring}, to avoid noisy estimates in bins with low occupancy, we found this approach to produce systematically higher baselines. We therefore used the equal-width bins and posterior predicted bin occupancies throughout for our baselines.

\begin{wrapfigure}{r}{0.42\textwidth}
    \centering
    \vspace{-10pt}
    \includegraphics[width=0.40\textwidth]{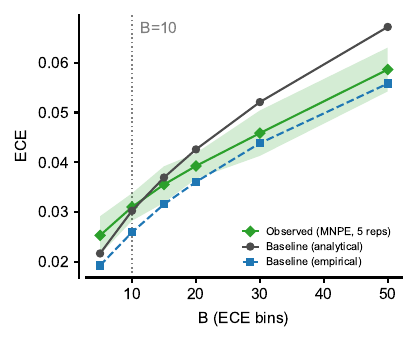}
    \caption{Observed ECE and its two baselines against the bin count $B$
    (queueing model, $N = 1{,}000$, $K = 5$, five training runs).}
    \label{fig:ece_bins}
    \vspace{-10pt}
\end{wrapfigure}

It is worth noting that both empirical baselines for ECE and EoD depend on the number of calibration samples $N$ and the number of bins $B$, which should match the number used to evaluate the calibration of the posterior. 
For the discrete calibration we used $B = 10$ equal-width bins tiling the interval $[0,1]$ in all experiments.
Both the observed ECE and its baseline grow with $B$, since more bins leave fewer test cases per bin, and both scale as $\sqrt{B/N}$, so that the two move together (\Figref{fig:ece_bins}).
On the queueing model at $N = 1{,}000$ and $K = 5$, the observed ECE rises from $0.025$ at $B = 5$ to $0.059$ at $B = 50$, while the empirical baseline rises from $0.019$ to $0.056$.
None of our conclusions depend on the choice of $B = 10$: the observed ECE stays inside its $95\%$ finite-sample null interval at every bin count we examined.

Since computing calibration requires $N$ simulations $\vx \sim p(\vx \mid \vtheta)$, $N$ is typically limited to a few hundred (see Appendix~\ref{app:exp_details} for the values used in each experiment).
At the same time, increasing $N$ yields tighter baselines; in the limit $N \to \infty$, the baseline converges to zero, implying that any approximate posterior will exhibit some deviation from perfect calibration. 

\subsection{Alternative Calibration checks} \label{app:alternative_calibration}
Instead of investigating the calibration of each marginal distribution individually, other functions can be used that project into one dimension \citep{talts2018validating}. The canonical projection using the posterior log-probability \citep{deistler2022truncated}, however, fails in the mixed-parameter setting. The theoretical guarantee that rank statistics are uniform under a well-calibrated posterior relies on the posterior being absolute continuous \citep{talts2018validating}. This condition is violated by discrete components. 
In practice, the discrete components tend to dominate the joint log-probability, masking miscalibration in the continuous parameters. 
An alternative approach is to separate the posterior into its discrete and continuous components. For the continuous parameters we can then use the log-probability on the ``joint marginal'' of all continuous dimensions $p(\vtheta_c \mid \vx)$ as projection function while still applying marginal calibration to the discrete dimensions as before. 
However, to be consistent across dimensions $i,j$, we assess calibration of the marginals of $\vtheta_c^i$ and $\vtheta_d^j$ separately as described in Section~\ref{subsec:calibration}.

\subsection{Conditional calibration of the continuous parameters}
\label{app:conditional}

The calibration checks of Section~\ref{subsec:calibration} assess each marginal separately, and in the mixed setting this can hide miscalibration of the coupling between the discrete and continuous factors.
The continuous marginal is a mixture over discrete states, $q(\vtheta_c \mid \vx) = \sum_{d} q(\vtheta_d = d \mid \vx)\, q(\vtheta_c \mid \vtheta_d = d, \vx)$, so marginal ranks average over states: continuous mass placed in the wrong discrete state, or conditional errors of opposite sign in different states, can cancel and leave the marginal ranks uniform.
We therefore test the conditionals directly by running SBC within discrete states.

For each of $1{,}000$ held-out pairs $(\vtheta^*, \vx)$ we fix the discrete parameters to their true value $\vtheta_d^*$, draw samples from $q(\vtheta_c \mid \vtheta_d^*, \vx)$, and rank the true $\vtheta_c^*$ among them.
Each run is summarized by the Kolmogorov--Smirnov statistic of these ranks, whose expectation under perfect calibration at $N = 1{,}000$ is $0.027$, with $0.043$ as the $95\%$ level.
On the queueing model, the conditional statistic for the arrival rate $\gamma$ is $0.044$ in the median over five training runs (range $0.034$--$0.055$), compared to $0.042$ ($0.023$--$0.057$) for marginal SBC and $0.034$ ($0.022$--$0.035$) for the joint log-probability projection of Appendix~\ref{app:alternative_calibration}.
The conditional statistic exceeds the marginal one in four of five runs, indicating a small residual miscalibration of the coupling for $\gamma$, at the edge of what $N = 1{,}000$ test cases resolve.

On the scaling model of Appendix~\ref{app:scaling} the discrete states are per-dimension, and grouping the SBC ranks of $\theta_i$ by the true value of $z_i$ gives $2D$ groups.
Every group passes at $D = 2$, one or two of the sixteen deviate at $D = 8$, and at $D = 32$ a minority of dimensions deviates clearly while the majority remain calibrated, at all training budgets.

Since calibration is a necessary rather than a sufficient condition, and the sampler never conditions on the true discrete state, we additionally check the sampled conditionals for accuracy where a reference posterior is available.
On the queueing model, we split the model's ancestral samples at $\vx_o$ by the discrete state was drawn with and compare each group against the reference conditional for that state.
The thirteen configurations with sufficient samples in every training run carry $99.7\%$ of the reference posterior mass, and their per-configuration C2ST of $0.553 \pm 0.007$ (range $0.52$--$0.62$) matches the model's joint C2ST of $0.535 \pm 0.006$, with no configuration-specific failure.

In both models, conditional calibration degrades before the marginal diagnostics do, and we therefore recommend it alongside the marginal checks.

\section{Additional experiments and ablations}
\label{app:ablations}

Unless stated otherwise, results in this appendix are means over three training seeds and uncertainties are standard errors of the mean.
Queueing-model results use the tuned architecture of Appendix~\ref{app:exp_details} at $N = 10^5$ simulations, the $1{,}000$-case held-out test set, and the quadrature reference of Section~\ref{sec:queue}; results on the Gaussian example are compared against its analytic posterior.

\subsection{Scaling in the size of the discrete space}
\label{app:scaling}

MNPE does not enumerate the discrete space, so its cost is independent of the number of configurations. 
To see how accuracy behaves as that space grows, we scale up the Gaussian example of Section~\ref{sec:toy}: the model becomes $D$ independent copies of the same one-dimensional problem, with the same class offset and observation noise, and in a second sweep $D = 2$ copies whose discrete parameter takes $K$ equally spaced values instead of two. 
The analytical posterior factorizes over dimensions, so it remains available in closed form at every $D$, and it holds the per-dimension difficulty fixed, so the sweeps isolate the size of the discrete space from the difficulty of each dimension. The queueing model covers the case of \emph{dependent} discrete parameters, at $D = 2$.

\begin{figure}[h]
    \centering
    \includegraphics[width=\textwidth]{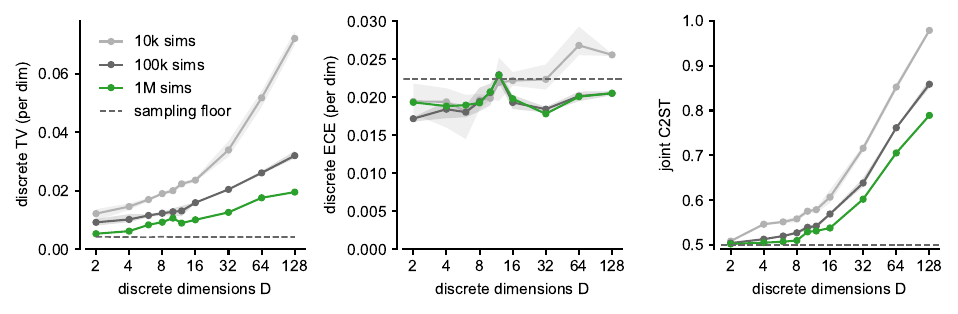}
    \caption{\textbf{Scaling in the number of discrete dimensions.} Per-dimension
    total variation of the discrete posterior against the exact posterior
    \textbf{(a)}, per-dimension discrete ECE \textbf{(b)} and joint C2ST
    \textbf{(c)}, as a function of $D$, one curve per simulation budget. Bands
    show the range over three training seeds ($10^6$: single seed). Dashed:
    the Monte Carlo measurement floor for total variation at $M = 4{,}000$
    posterior samples \textbf{(a)}, the finite-sample ECE baseline of
    Appendix~\ref{app:ece_baseline} \textbf{(b)}, and the C2ST chance level
    \textbf{(c)}.}
    \label{fig:scaling}
\end{figure}

Degradation with $D$ is monotone, smooth for different budgets (\Figref{fig:scaling}a). 
Per-dimension total variation at $D = 128$ falls from $0.072$ to $0.032$ to $0.020$ as the budget rises from $10^4$ to $10^6$, and grows roughly as $D^{0.3}$ at fixed budget; at the other end, $D = 2$ with $10^6$ simulations reaches $0.005$, which is the $0.004$ floor imposed by the total variation of $4{,}000$ posterior samples (dashed). 
Per-dimension calibration stays at or below its finite-sample baseline at the two larger budgets, and rises slightly above it only at $10^4$ simulations with $D \ge 64$ (\Figref{fig:scaling}b). 
The joint C2ST rises with $D$ ($0.86$ at $D = 128$, $10^5$ simulations), but note that a joint C2ST score over $256$ dimensions tends to be sensitive to slightest deviations in individual dimensions, even when each dimension is individually accurate.

\begin{figure}[h]
    \centering
    \includegraphics[width=\textwidth]{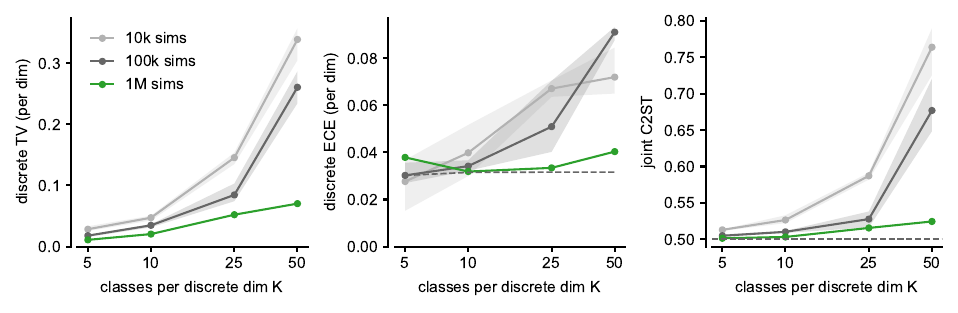}
    \caption{\textbf{Scaling in the number of classes per discrete dimension.}
    As \Figref{fig:scaling}, for $D = 2$ dimensions with $K$ classes each and the network width held fixed, so that the sweep isolates the class count from
    capacity. 
    Panel \textbf{(b)}'s baseline is class-count aware.}
    \label{fig:ksweep}
\end{figure}

When class count is increased we find that per-dimension total variation at $K = 50$ is $0.261$ at $10^5$ simulations, and the joint C2ST reaches $0.68$ (\Figref{fig:ksweep}). Calibration degrades in step with accuracy rather than independently. 
Training budget and capacity are the main limitations. At fixed width, total variation at $K = 50$ falls to $0.070$ at $10^6$ simulations and the joint C2ST returns to $0.52$. Quadrupling the width at fixed budget roughly halves the error. 

In addition to the marginal checks performed here, we also performed a conditional calibration check of the coupling between each discrete dimension and its continuous partner (Appendix~\ref{app:conditional}). We found that the per-dimension marginals stay accurate past the point, $D \ge 8$, where some conditionals begin to degrade.

Note all results use the discrete head of Section~\ref{sec:methods}, a masked multilayer perceptron, which depends on $D$. 
However, MNPE places no constraint on the discrete architecture and weight sharing architectures (e.g., masked attention over the dimensions, or a permutation-equivariant set encoder where dimensions are exchangeable) may scale better. We leave this to future work.

\subsection{Joint versus separate training of the two heads}
\label{app:jointsep}
  
The two heads of Section~\ref{sec:methods} are trained jointly. 
Although the loss decomposes exactly and the subnetworks share no parameters, the procedure itself still couples them, because the optimizer does not apply raw gradients: gradient-norm clipping is global, so the discrete head's updates are rescaled by a factor set almost entirely by the continuous head, and checkpoint selection returns both heads from the single epoch that minimizes the \emph{summed} validation loss. 
To quantify the effect of these couplings, we also train each head under its own loss term, which removes both. 
The two schemes agree to $0.0001$ in mean joint C2ST and to $0.0008$ in mean ECE, within the seed-to-seed spread of either. 
The only difference is training cost: the separately trained discrete head converges in about half the epochs
($50.0 \pm 6.7$ against $94.7 \pm 9.0$), no longer held open by the continuous
head's validation curve. 
Note this equivalence is specific to a parameter-free condition embedding. A learned embedding of $\vx$ shared by both heads would couple the gradients directly, and the choice between joint and separate training would have to be revisited.

\subsection{Standard NPE on dequantized discrete parameters}
\label{app:dequant}

An alternative to a dedicated discrete head is to assume only continuous parameters and use standard NPE. To construct such a baseline, we add uniform noise, $\tilde\theta_d = \theta_d + u$ with $u \sim \mathcal{U}[0, 1)$, train NPE on the resulting continuous parameter vector, and recover discrete values by flooring, taking categorical probabilities as sample fractions per unit interval. 
Flow family, hyperparameters, budgets, training-data seeds and evaluation are those of MNPE, so the comparison isolates the treatment of the discrete parameters.

\begin{table}[h]
    \centering
    \caption{MNPE against NPE on uniformly dequantized discrete parameters, at
    matched budget, architecture and evaluation; mean $\pm$ s.e.m.\ over three
    training seeds (five training runs for MNPE on the queueing model). Queueing:
    $N = 10^5$, C2ST against the quadrature reference (floor $0.502$). Gaussian:
    $N = 10^4$, against the analytic posterior; the MNPE values are the single
    pretrained model of Section~\ref{sec:toy}.}
    \begin{tabular}{lccc}
        \toprule
        & \multicolumn{2}{c}{C2ST, joint} & error in $p(z = 1)$ \\
        \cmidrule(lr){2-3}\cmidrule(lr){4-4}
        Method & Gaussian & Queueing & Gaussian \\
        \midrule
        MNPE & $0.498$ & $0.536 \pm 0.006$ & $0.006$ \\
        Dequantized NPE & $0.505 \pm 0.001$ & $0.630 \pm 0.008$ & $0.009 \pm 0.002$ \\
        \bottomrule
    \end{tabular}
    \label{tab:dequant}
\end{table}

On the single binary parameter of the Gaussian example the baseline is close to MNPE but worse on each metric we measure (Table~\ref{tab:dequant}).
On the queueing model, with two five-valued parameters, the gap is larger: a joint C2ST
of $0.630$ against MNPE's $0.536$.
This gap is stable across a change of flow family and two independent evaluation redraws.

Three structural properties of dequantization hold regardless of budget or architecture. 
It imposes an ordinal adjacency on the categories, which suits server counts but not nominal categories such as the channel identities of the Hodgkin--Huxley example; it yields sample fractions rather than exact categorical probabilities; and probability mass outside the valid interval, $0.6$--$1.3\%$ here, has to be removed by rejection. 

\subsection{Factorization of the joint posterior}
\label{app:factorization}

The discrete-first factorization of Section~\ref{sec:methods} has two natural alternatives.
First, reversing the direction, $q(\vtheta_c \mid \vx)\,q(\vtheta_d \mid \vtheta_c, \vx)$, is an equally valid factorization.
Second, dropping the conditioning, $q(\vtheta_d \mid \vx)\,q(\vtheta_c \mid \vx)$, to remove the dependence between discrete and continuous parameters.
For the two ablations, we implemented the discrete stages as stand-alone classifiers trained by cross entropy on simulator pairs. The reversed and independent arms share one continuous marginal flow and differ only in whether they are used by the discrete stage.

\begin{table}[h]
    \centering
    \caption{Joint C2ST for three factorizations of the mixed posterior, mean
    $\pm$ s.e.m.\ over three training seeds (five for the MNPE queueing arm).
    Gaussian example: $N = 10^4$, against the analytic posterior; the MNPE value
    is the single pretrained model of Section~\ref{sec:toy}, so no s.e.m.\ is
    available. Queueing model: $N = 10^5$, against the quadrature reference.}
    \begin{tabular}{lcc}
        \toprule
        Factorization & Gaussian & Queueing \\
        \midrule
        $q(\vtheta_d \mid \vx)\,q(\vtheta_c \mid \vtheta_d, \vx)$ (MNPE) & $0.498$ & $0.535 \pm 0.005$ \\
        $q(\vtheta_c \mid \vx)\,q(\vtheta_d \mid \vtheta_c, \vx)$ (reversed) & $0.506 \pm 0.001$ & $0.572 \pm 0.002$ \\
        $q(\vtheta_d \mid \vx)\,q(\vtheta_c \mid \vx)$ (independent) & $0.575 \pm 0.001$ & $0.877 \pm 0.002$ \\
        \bottomrule
    \end{tabular}
    \label{tab:factorization}
\end{table}

When the conditioning is dropped MNPE fails to predict the joint correctly (Table~\ref{tab:factorization}). The independent arm estimates $p(z = 1)$ to $0.008$ in the Gaussian example while its conditional means are off by $0.80$, and its joint C2ST rises to $0.575$, peaking at $0.65$ at the observation where the two posterior modes are most balanced.
On the queueing model it reaches $0.877$.
When the direction is reversed the difference is most visible in sample efficiency: the reversed arm comes within $0.01$ of the forward one at $10^4$ simulations on the Gaussian example, but reaches only $\approx 0.60$ at $10^3$, where the flow misplaces mass in the bimodal continuous marginal it has to fit first, and it remains $0.037$ above forward on the queueing model.
The discrete-first order avoids this, since each per-state conditional is unimodal.

\subsection{Propagation of discrete error into the continuous posterior}
\label{app:propagation}

Because MNPE samples ancestrally, the continuous posterior returns a mixture over discrete states, $q(\vtheta_c \mid \vx) = \sum_d q(d \mid \vx)\, q(\vtheta_c \mid d, \vx)$, so an error in the discrete head re-weights the components.
Splitting the difference from the true posterior into a wrong-weights and a wrong-components term and applying the triangle inequality to the total variation distance (TV) gives
\begin{equation*}
    \mathrm{TV}\big(q(\vtheta_c \mid \vx),\, p(\vtheta_c \mid \vx)\big)
    \;\le\; \mathrm{TV}\big(q(\vtheta_d \mid \vx),\, p(\vtheta_d \mid \vx)\big)
    \;+\; \max_d \mathrm{TV}\big(q(\vtheta_c \mid d, \vx),\, p(\vtheta_c \mid d, \vx)\big),
\end{equation*}
so discrete error enters at most linearly, with coefficient one, and cannot be amplified by the continuous stage.

\begin{wrapfigure}{r}{0.5\textwidth}
    \centering
    \vspace{-5pt}
    \includegraphics[width=0.48\textwidth]{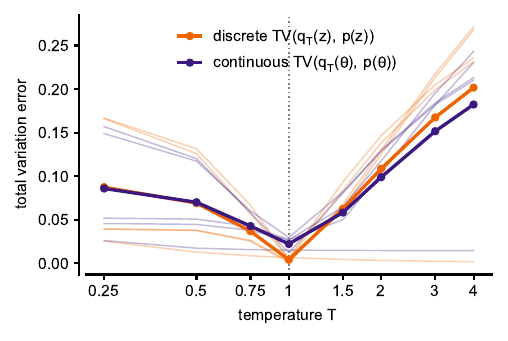}
    \caption{\textbf{Discrete error does not amplify downstream.} Total variation error
    of the discrete posterior (orange) and of the resulting continuous posterior
    (purple) as a function of the temperature $T$; thin lines are the five test
    observations, bold lines their mean, and the dotted line marks the
    undistorted head at $T = 1$.}
    \label{fig:propagation}
    \vspace{-10pt}
\end{wrapfigure}

We confirm this on the Gaussian example by deliberately distorting the sampling from the trained discrete head with temperature scaling, $q_T(z \mid \vx) \propto q(z \mid \vx)^{1/T}$, while holding the continuous head fixed (\Figref{fig:propagation}).
Across injected discrete errors up to $\mathrm{TV} = 0.27$, the TV bound growing at most linearly holds at every setting, and the excess continuous error grows with a slope of $0.78$ relative to the excess discrete error ($0.50$ to $0.86$ across observations).

\subsection{Ordering of the discrete variables}
\label{app:ordering}

For $D$ discrete parameters the autoregressive factorization of the discrete MNPE head admits $D!$ possible orderings, all exact factorizations of the same joint mass function.
We test whether this ordering matters on the queueing model.
For this purpose we compare the two possible orderings and use the same training data, initialization, batch ordering and continuous head, such that only the ordering differs.
As the primary comparison we use the held-out log mass of the discrete factor on the $1{,}000$-case test set, paired per test case.

\begin{table}[h]
    \centering
    \caption{Sensitivity to the ordering of the discrete variables in the
    autoregressive factorization, queueing model at $N = 10^5$, mean $\pm$
    s.e.m.\ over three seeds. The ECE null interval is $\approx 0.03$ wide at $N = 1{,}000$ (Appendix~\ref{app:ece_baseline}), so that column can only detect a large
    effect.}
    \begin{tabular}{lcccc}
        \toprule
        Ordering & held-out $\log q(\vtheta_d \mid \vx)$ & joint C2ST & discrete TV & ECE ($c_1$ / $c_2$) \\
        \midrule
        $c_1 \rightarrow c_2$ & $-2.6018 \pm 0.0011$ & $0.534 \pm 0.003$ & $0.037 \pm 0.006$ & $0.026 / 0.035$ \\
        $c_2 \rightarrow c_1$ & $-2.6051 \pm 0.0007$ & $0.532 \pm 0.004$ & $0.036 \pm 0.010$ & $0.029 / 0.034$ \\
        \bottomrule
    \end{tabular}
    \label{tab:ordering}
\end{table}

The ordering swap changes held-out log mass by only $0.0033$ nats ($0.13\%$ of the mean), an effect too small to resolve on this test set (paired $|z| \le 1.5$ in every seed).
Every other metric differs by less than its seed-to-seed spread (Table~\ref{tab:ordering}), and per-dimension ECE lies inside its $95\%$ finite-sample null interval (Appendix~\ref{app:ece_baseline}) for both orderings, both dimensions and all seeds.
The two orderings do reach different solutions, their mutual discrete total variation is $0.028$, but neither is closer to the reference.
Their one systematic difference is training cost: the swapped ordering needs about $40\%$ more epochs to converge, with non-overlapping ranges across seeds.

\section{Additional examples}

\subsection{Coal mining disaster changepoint inference}
\label{app:coal}

The coal mining switchpoint simulator is presented in the PyMC
documentation\footnote{\url{https://www.pymc.io/projects/examples/en/latest/howto/marginalizing-models.html}}
as an example for dealing with discrete model parameters using automatic
marginalization. We include it here as a second direct comparison between amortized MNPE
and MCMC inference. This example differs from the queueing model in two key respects:
the observation is high-dimensional ($x \in \mathbb{N}^{111}$, one count per year),
requiring an embedding network, and the discrete space is substantially larger
($|\mathcal{D}| = 111$ switchpoints versus $|\mathcal{D}| = 25$ server configurations).
At the same time, this inference problem is arguably easier: the switchpoint only
determines where the rate changes, inducing weaker discrete--continuous coupling than
the queueing simulator's stability constraint, and the posterior is unimodal.

The coal mining disaster dataset records the number of coal mining disasters per year in
the United Kingdom from 1851 to 1961 \citep{jarrett_1979}. A visible decline in disaster
frequency around 1890, attributed to improved safety regulations, makes this a classic
benchmark for Bayesian changepoint analysis \citep{raftery_akman_1986}. While a
continuous relaxation of the switchpoint is possible for this model, we keep the
discrete formulation as a representative test case for the broader class of models with
inherently discrete parameters.

The observed disaster count $y_t$ in year $t$ follows a Poisson likelihood with a rate
that switches at an unknown year~$s$
\begin{align*}
\begin{split}
    s &\sim \mathrm{DiscreteUniform}(1851,\, 1961), \\
    \lambda_{\mathrm{early}},\, \lambda_{\mathrm{late}} &\sim \mathrm{Exponential}(1), \\
    y_t \mid s, \lambda_{\mathrm{early}}, \lambda_{\mathrm{late}} &\sim \mathrm{Poisson}\!\bigl(\lambda_{\mathrm{early}}\,[t < s] + \lambda_{\mathrm{late}}\, [t \geq s]\bigr).
\end{split}
\end{align*}
In the notation of \Secref{sec:methods}, the discrete parameter is $\theta_d = s \in
\{1851, \ldots, 1961\}$ ($|\mathcal{D}| = 111$) and the continuous parameters are
$\theta_c = (\lambda_{\mathrm{early}}, \lambda_{\mathrm{late}}) \in \mathbb{R}_{>0}^2$.
The observation is the full sequence of annual disaster counts $x = (y_{1851}, \ldots,
y_{1961}) \in \mathbb{N}^{111}$.

\begin{figure}[t]
    \centering
    \includegraphics[width=\textwidth]{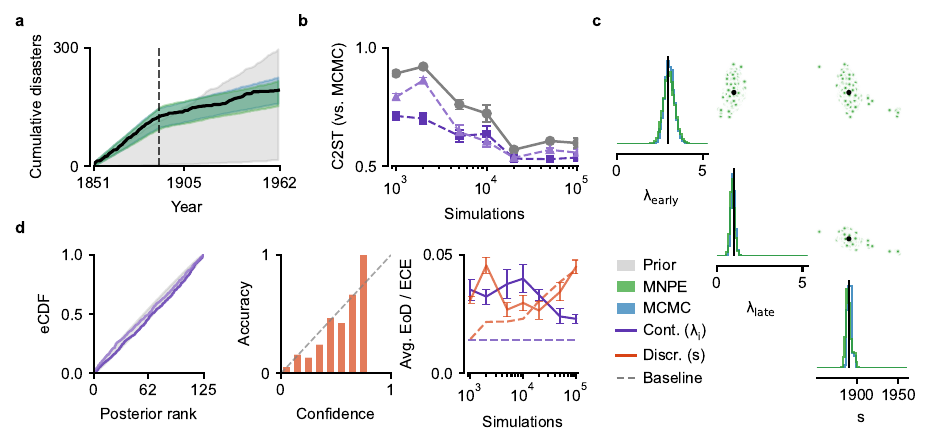}
    \caption{\textbf{Coal mining changepoint simulator} ($|\mathcal{D}| = 111$).
    \textbf{(a)}~Cumulative coal mining disasters in the UK (1851--1961, black). The change in slope around 1890 (vertical line) reflects improved safety regulations. Predictive distributions (5--95th percentile) are shown for prior (grey), MNPE (green) and MCMC reference (blue). 
    \textbf{(b)}~C2ST versus training budget for joint and per-parameter comparisons against the MCMC reference. Error bars show $\pm 1$ SEM over five independent training runs.
    \textbf{(c)}~Joint posterior over all three parameters: MNPE (green) versus marginalized MCMC (blue). Black markers indicate historical estimates.
    \textbf{(d)}~Calibration diagnostics. \emph{Left:}~SBC rank eCDF for the continuous parameters ($\lambda_{\mathrm{early}}$, $\lambda_{\mathrm{late}}$) at $10^5$ simulations, with 95\% uniform confidence band in grey. \emph{Center:} ECE reliability diagram for the discrete switchpoint at $10^5$ simulations. \emph{Right:}~Expected calibration error for continuous (EoD, violet) and discrete (ECE, orange) parameters, averaged across five seeds and all marginals across budgets. Dashed lines indicate expected optimal calibration error for the given training budget.
    }
    \label{fig:coal}
\end{figure}

The discrete switchpoint is analytically marginalized reducing inference to a purely
continuous problem amenable to NUTS \citep{pymc_2023,hoffman_gelman_2014}. The discrete
posterior $p(s \mid x)$ is then recovered by Rao-Blackwellization: for each posterior
draw of $(\lambda_{\mathrm{early}}, \lambda_{\mathrm{late}})$, the conditional $p(s \mid
\lambda_{\mathrm{early}}, \lambda_{\mathrm{late}}, x)$ is computed analytically via the
Poisson log-likelihood summed over all years. The marginalization cost scales as
$O(|\mathcal{D}|)$ per MCMC step, requiring 111 Poisson log-likelihood evaluations per
gradient computation. While tractable for this simulator, this cost becomes prohibitive
for larger discrete spaces.

We performed a hyperparameter search to select an optimal MNPE architecture via Optuna
~\citep{akiba_optuna_2019}, optimizing the negative log-probability of the current
posterior estimate (NLTP) on a held-out validation set
following~\citet{lueckmann2021benchmarking}. The tuned architecture uses a neural spline
flow with 2~coupling transforms, 1~hidden layer of 64~units, and 10~rational-quadratic
spline bins. Since the observation $x \in \mathbb{N}^{111}$ is high-dimensional relative
to the three-dimensional parameter space, a fully-connected embedding network (1~hidden
layer of 64~units, output dimension~32) compresses the observation before the density
estimator. We also applied a variance-stabilizing $\sqrt{\cdot}$ transform to the
Poisson counts before training, which produces a more uniform variance across rate
regimes and improves z-scoring during training. We evaluate MNPE across seven training
budgets from $10^3$ to $10^5$ simulations, with five independent training runs per
budget.

The cumulative disaster count exhibits a clear change in slope around 1890, consistent
with the historical switchpoint. Posterior predictive cumulative curves from MNPE and
MCMC both tightly envelop the observed trajectory, whereas the prior predictive band
spans a much wider range (Fig.~\ref{fig:coal}a). With increasing training budget, the
gap between MNPE and MCMC narrows: joint C2ST falls to $0.56 \pm 0.01$ at $10^5$
simulations, and the marginal C2ST for both rate parameters approaches the chance level
of~$0.5$ (Fig.~\ref{fig:coal}b). Examining a single observation $\vx_o$ confirms these results: the MNPE posterior concentrates $\lambda_{\mathrm{early}}$ around
3.0~disasters per year and $\lambda_{\mathrm{late}}$ around~0.9, with the switchpoint
peaking near 1890, all closely matching the marginalized MCMC reference
(Fig.~\ref{fig:coal}c). SBC rank distributions at $10^5$ simulations show no systematic
deviation from uniformity for either continuous parameter (Fig.~\ref{fig:coal}d, left),
and the ECE reliability diagram for the 111-class switchpoint indicates well-calibrated
discrete predictions (Fig.~\ref{fig:coal}d, center). Across training budgets, continuous
calibration steadily improves toward the finite-sample baseline, while discrete
calibration remains close to the expected calibration of a perfectly calibrated classifier (Fig.~\ref{fig:coal}d, right).

\subsection{Hodgkin--Huxley simulator}
Below we show the full posterior distributions from Fig.~\ref{fig:hh}.

\begin{figure}[h]
    \centering
    \includegraphics{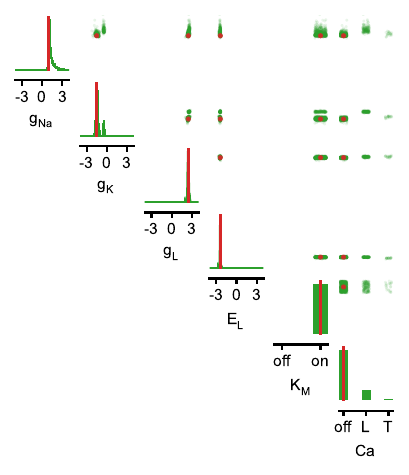}
    \caption{\textbf{Hodgkin-Huxley posterior.} 
    One and two dimensional posterior marginals of the full parameter space (four continuous and two discrete dimensions) for the observation shown in Fig.~\ref{fig:hh}a.
    }
    \label{app:fig:HH_posterior}
\end{figure}

\begin{figure}[t]
    \centering
    \includegraphics{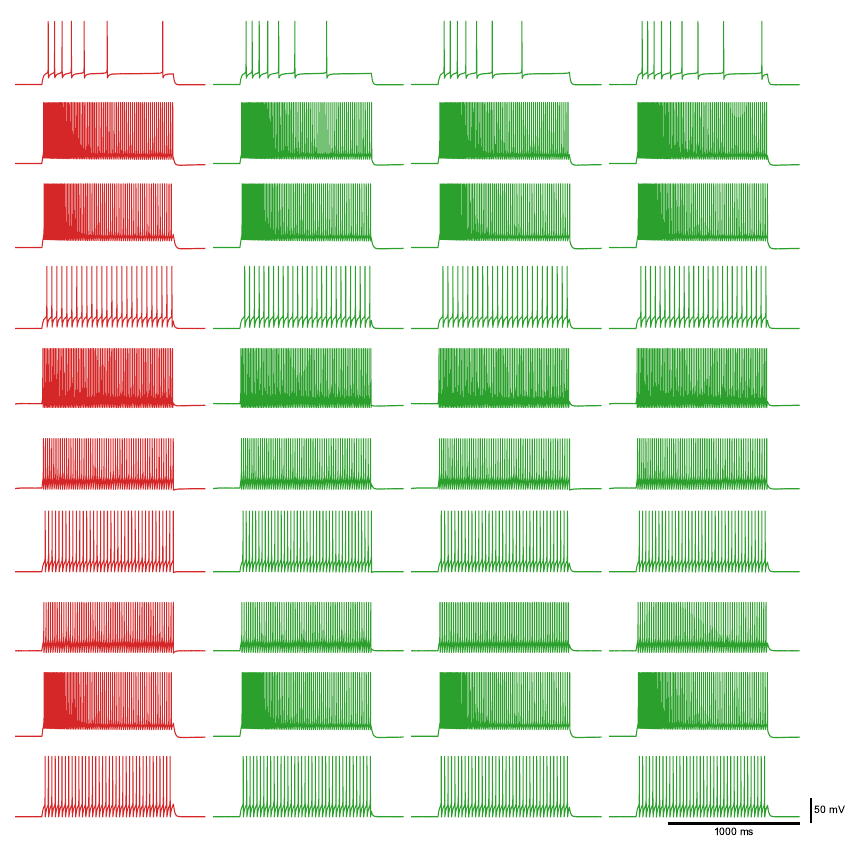}
    \caption{\textbf{Posterior predictives for the Hodgkin-Huxley simulator.} 
    Three posterior predictive samples (green) for 10 observations (red).
    }
    \label{app:fig:HH_example_traces}
\end{figure}

\section{Experimental details} \label{app:exp_details}

MNPE is integrated in the \href{https://sbi.readthedocs.io/en/stable/}{\texttt{sbi}} toolbox \citep{boeltsdeistler2025reloaded}, with the method implementation available \href{https://github.com/sbi-dev/sbi/blob/main/sbi/inference/trainers/npe/mnpe.py}{here}.
Code to reproduce all experiments and figures in this paper is available at \url{https://github.com/mackelab/mnpe-paper}.

\subsection{Gaussian example}

MNPE uses a neural spline flow with 5 coupling transforms and 4 blocks for the continuous parameters, and a MADE for the discrete parameters. Both networks have 3 hidden layers with 32 hidden features each. Training uses up to $N = 10{,}000$ simulations with learning rate set to $\eta=5e^{-4}$, validation set fraction of ten percent, and early stopping after 20 epochs without validation loss improvement. We repeat each training with five random seeds and average the evaluation results.

\subsection{Tandem queueing simulator}

The tandem queueing simulator consists of two M/M/$c$ queues in
series~\citep{gross_2008, jackson1957networks}, where M/M/$c$ denotes a queue
with memoryless (Poisson) arrivals, memoryless (exponential) service times, and
$c$~parallel servers. Customers arrive at station~1 with rate~$\gamma$, are
served by~$c_1$ parallel servers at rate~$\mu_1$, and then proceed to
station~2 with~$c_2$ servers at rate~$\mu_2$. The continuous parameters are the
arrival and service rates $\vtheta_c = (\gamma, \mu_1, \mu_2) \in
\mathbb{R}_{>0}^3$ and the discrete parameters are the server counts
$\vtheta_d = (c_1, c_2) \in \{2, 3, 4, 5, 6\}^2$ ($|\mathcal{D}| = 25$).
The parameters are drawn from the prior
\begin{align*}
\begin{split}
    \gamma &\sim \mathrm{LogNormal}(\log 9,\; 0.3), \\
    \mu_1 &\sim \mathrm{LogNormal}(\log 8,\; 0.3), \\
    \mu_2 &\sim \mathrm{LogNormal}(\log 5,\; 0.3), \\
    c_1, c_2 &\sim \mathrm{DiscreteUniform}\{2, 3, 4, 5, 6\}.
\end{split}
\end{align*}

The per-station traffic intensity is $\rho_i = \gamma / (c_i \mu_i)$ and both
stations must satisfy $\rho_i < 1$ for the system to be stable. In steady
state, the expected number of customers waiting in the queue at station~$i$ is
given by the standard M/M/$c$ result \citep{gross_2008}
\begin{equation*}
    \mathbb{E}[Q_i]
    \;=\; \frac{r_i^{\,c_i}\,\rho_i}{c_i!\,(1 - \rho_i)^2}\;\pi_{0,i}\,,
    \qquad\text{where}\quad
    \pi_{0,i} = \left[\sum_{n=0}^{c_i - 1}\frac{r_i^{\,n}}{n!}
    \;+\; \frac{r_i^{\,c_i}}{c_i!\,(1 - \rho_i)}\right]^{-1},
\end{equation*}
with offered load $r_i = \gamma / \mu_i$. By Jackson's theorem, the two
stations are independent in steady state, so the joint observation likelihood
factorizes across stations. This simulator exhibits strong discrete--continuous
coupling: expected queue lengths grow steeply as $\rho_i \to 1$, making the
posterior sensitive to the interaction between server counts and service rates.

Given a time horizon $T = 100$, the simulator produces five observations:
arrival counts, completion counts at each station, and two queue lengths. The
three count dimensions follow Poisson distributions with rate~$\gamma T$
\begin{align*}
    n_{\mathrm{arr}} &\sim \mathrm{Poisson}(\gamma T), &
    n_{\mathrm{comp},1} &\sim \mathrm{Poisson}(\gamma T), &
    n_{\mathrm{comp},2} &\sim \mathrm{Poisson}(\gamma T).
\end{align*}
In steady state, the throughput at each station equals the arrival rate, so all
three counts share the same rate and identify only~$\gamma$. The two queue
length observations provide the information needed to identify the service
rates~$\mu_1$ and~$\mu_2$, since $\mathbb{E}[Q_i]$ depends nonlinearly on
$\rho_i = \gamma / (c_i \mu_i)$
\begin{equation*}
    q_i \;\sim\; \mathrm{TruncatedNormal}\!\bigl(\mathbb{E}[Q_i],\;
    \sigma_{\mathrm{obs}},\; \mathrm{lower}{=}0\bigr),
    \qquad \sigma_{\mathrm{obs}} = 0.1.
\end{equation*}
Prior draws are discarded unless both stations satisfy $\rho_i < 0.999$, a small margin below the stability boundary that avoids the numerical pathology of $\mathbb{E}[Q_i]$ as $\rho_i \to 1$. 
This removes approximately $13\%$ of prior samples.

Since NUTS cannot directly handle discrete parameters, a reference must marginalize over the 25 server-count configurations. 
Marginalizing analytically leaves a multimodal continuous posterior on which NUTS does not converge: the configurations induce well-separated modes in $(\mu_1, \mu_2)$ that chains cannot cross. Running one chain per configuration with a stability guard $\rho_i < 0.999$ converges reliably, but combining configurations then requires their marginal likelihoods, and estimating these by a Laplace approximation leaves an error shared by any implementation that uses it (PyMC and NumPyro).

We therefore compute the reference deterministically. 
Conditional on the arrival rate~$\gamma$ the two stations are independent, which reduces each configuration's evidence to nested one-dimensional integrals. 
We evaluate these by quadrature and sample by rejection, so neither a Markov chain nor an approximation of the configuration weights is involved. 
As an additional independent check we marginalize by explicit log-sum-exp and sample the three-dimensional target with sequential Monte Carlo, which agrees closely with the quadrature approach (C2ST $0.515$, discrete total variation $0.004$); per-configuration NUTS references in PyMC~\citep{pymc_2023} and NumPyro~\citep{phan2019_numpyro} are further away (C2ST $0.52$--$0.53$). 
To estimate the C2ST we used $10{,}000$ samples, a random forest classifier and 5-fold cross-validation.

The MNPE inference network consists of a categorical MADE for the discrete parameters and a neural spline flow (NSF) for the continuous parameters. Both components share the same width and depth: 4~hidden layers of 256~units. The NSF uses 4~coupling transforms with 9~rational-quadratic spline bins. Because queue length observations exhibit extreme right skew (median ${\sim}0.1$, tail extending to ${\sim}10$), we apply a $\log(1{+}x)$ transform to the two queue length dimensions before training. Training data is generated from the joint prior defined above, ensuring that the simulated data distribution matches the likelihood used for the reference posterior. 
Training uses $N = 1{,}000$ to $100{,}000$ simulations with a batch size of 256, a learning rate of $\eta = 1.1e^{-4}$, a validation set fraction of ten percent, and early stopping after 20 epochs without validation loss improvement.
We repeat each training with five random seeds and average the evaluation results. Calibration is assessed on a held-out test set of $N = 1{,}000$ samples, drawing $1{,}000$ posterior samples per observation.

\begin{table}[h]
    \centering
    \caption{Wall-clock cost for the queueing model (Apple M3 Pro, CPU only).
    MNPE has a one-off simulation and training cost and then draws
    $10{,}000$ posterior samples per observation.
    Every alternative requires the closed-form likelihood and runs in full for
    each observation; the quadrature reference additionally exploits the
    conditional independence of the two stations, and the NUTS range spans a
    practical setting (two chains, $23/25$ configurations converged) to the
    reference setting (four chains, all converged).}
    \begin{tabular}{lr}
        \toprule
        & wall clock \\
        \midrule
        MNPE, one-off ($N = 10^4$ simulations) & $59$\,s \\
        MNPE, one-off ($N = 10^5$ simulations) & $493$\,s \\
        \midrule
        MNPE, per observation & $0.09$\,s \\
        SMC (analytic marginalization), per observation & $3.4$\,s \\
        Per-configuration NUTS, per observation & $42$--$90$\,s \\
        Quadrature reference, per observation & $82$\,s \\
        \bottomrule
    \end{tabular}
    \label{tab:runtime}
\end{table}

\subsection{Hodgkin--Huxley simulator}
The Hodgkin--Huxley simulator \cite{hodgkin1952quantitative} describes action potential generation in neural tissue by considering the change in cell's membrane potential $V(t)$ in response to an external input $I_{\text{in}}(t)$ and how different voltage activated ion channels open and close in response to it.

We consider sodium, potassium and leak currents to generate spiking, a slow non-inactivating $K^+$ current that allows for spike-frequency adaptation, and low- (T-type) and high-threshold (L-type) $Ca^{2+}$ currents \cite{pospischil2008minimal}

\begin{align*}
\begin{split}
     C\ \frac{dV_t}{dt} =& I_t - g_{Na}m^3h(V_t - E_{Na}) - g_{K} n^4(V_t - E_{K}) - g_{leak}(V_t - E_{leak})\\
     &- g_{M} p(V_t - E_{Kt}) \\
     &- g_{CaT}q^2r(V_t - E_{Ca}) \\
     &- g_{CaL}q^2(V_t - E_{Ca}) \\
     & + \mathcal{N}(0,\sigma),
\end{split}
\end{align*}

with maximal conductances of the sodium, potassium, leak, adaptive potassium and calcium ion channels $g_i,\ i \in \{Na,K,leak,M,CaL,CaT\}$ and associated reversal potentials $E_i$. $C$ denotes the membrane capacitance, $I_t$ denotes the input current per unit area and $n, m, h, q, r$ and $p$ the fraction of opened channel gates. 
The gating dynamics can be expressed as

\begin{subequations}
\begin{align*}
         \frac{dz_t}{dt} &= \alpha_z(V_t)\,(1-z_t) - \beta_z(V_t) z_t \\
    \frac{dp_t}{dt} &= (p_{\infty}(V_t) - p_t)/\tau_p(V_t),
\end{align*}
\end{subequations}

with rate constants $\alpha_z(V_t)$ and $\beta_z(V_t)$ of $z \in \{m,n,h,q,r\}$ and $p$. For more detailed equations see \cite{pospischil2008minimal}.
The parameter values and bounds we used were adapted from \citep{pospischil2008minimal} and \cite{goncalves2020training} to cover biological meaningful ranges and are listed in Tab.~\ref{tab:HH_params}.

\begin{table}[h]
    \centering
    \caption{Parameter bounds and values that were used for our experiments.}
    \begin{tabular}{lrrrr}
        \toprule
        Parameter & Lower bound & Upper bound & Fixed Value&Observation Fig.\ref{fig:hh}a (rounded)\\
         \midrule
        $g_{Na}$ (mS) &8 & 80  & -& 60.4 \\
        $g_{K}$ (mS) & 1.5 & 15  & -& 3.9\\
        $E_{Na}$ (mV) & - & - & 50& 50\\ 
        $E_{K}$ (mV) & - & - & -90& -90\\
        $g_{leak}$ (mS) & 0.01& 0.1 & -&0.09\\
        $E_{leak}$ (mV) & -80 & -60 & -& -78.3\\
        $g_{M}$ (mS) & - & - & \{0, 0.03\}& 0.03\\
        $E_{Ca}$ (mV) & - & - & 120& 120\\
        $g_{CaL}$ (mS) & - & - & \{0, 0.1\}& 0\\
        $g_{CaT}$ (mS) & - & - & \{0, 0.4\} &  0\\
        $\tau_{p}$ (s) & - & - & 1& 1\\
        \bottomrule
    \end{tabular}
    \label{tab:HH_params}
\end{table}

We used a step current $I_{inj}$ of $2\mu A/cm^2$ for $1000ms$ and run the simulation for $1450ms$. This stimulus and recording protocol corresponds to the voltage recordings from the \citet{allen_database}.
Finally, we added independent Gaussian noise with a standard deviation of $\sigma = 0.1 mV$ to each time point to simulate the stochasticity of the underlying biophysical processes.

MNPE uses a neural spline flow with 5 coupling transforms and 8 blocks for the continuous parameters, and a MADE for the discrete parameters. 
Both networks have 6 hidden layers with 50 hidden features each. Training uses $N = 100, 500, 1000, \ldots, 100{,}000$ simulations with a batch size of 500. We repeat each training with three random seeds and average the evaluation results.

For the MSE evaluation, we use a test set of samples $(\vtheta_o, \vx_o)$ drawn from the joint $p(\vtheta)\,p(\vx \mid \vtheta)$. 
We then draw a posterior sample $\vtheta_{post} \sim p(\vtheta \mid \vx_o)$, run a simulation $\vx_{post} \sim p(\vx \mid \vtheta_{post})$, and compute the MSE between $\vx_o$ and $\vx_{post}$ for each test sample.
To assess the calibration of the continuous and discrete parameters we use a test set of $500$ samples $(\vtheta_o, \vx_o)$, drawing $1{,}000$ posterior samples for continuous SBC per observation~$\vx_o$.

\end{document}